\documentclass[journal]{elsarticle}

\usepackage{scalerel}
\usepackage{tikz}
\usetikzlibrary{svg.path}
\definecolor{orcidlogocol}{HTML}{A6CE39}
\tikzset{
    orcidlogo/.pic={
        \fill[orcidlogocol] svg{M256,128c0,70.7-57.3,128-128,128C57.3,256,0,198.7,0,128C0,57.3,57.3,0,128,0C198.7,0,256,57.3,256,128z};
        \fill[white] svg{M86.3,186.2H70.9V79.1h15.4v48.4V186.2z}
        svg{M108.9,79.1h41.6c39.6,0,57,28.3,57,53.6c0,27.5-21.5,53.6-56.8,53.6h-41.8V79.1z M124.3,172.4h24.5c34.9,0,42.9-26.5,42.9-39.7c0-21.5-13.7-39.7-43.7-39.7h-23.7V172.4z}
        svg{M88.7,56.8c0,5.5-4.5,10.1-10.1,10.1c-5.6,0-10.1-4.6-10.1-10.1c0-5.6,4.5-10.1,10.1-10.1C84.2,46.7,88.7,51.3,88.7,56.8z};
    }
}

\newcommand\orcidicon[1]{\href{https://orcid.org/#1}{\mbox{\scalerel*{
                \begin{tikzpicture}[yscale=-1,transform shape]
                \pic{orcidlogo};
                \end{tikzpicture}
            }{|}}}}

\usepackage[hidelinks]{hyperref}
\usepackage{lineno}
\modulolinenumbers[5]

\usepackage{amssymb}
\usepackage{graphicx}
\usepackage{subfigure}
\usepackage{amsmath}
\usepackage{adjustbox}
\usepackage{multirow}
\usepackage{mathtools}
\usepackage{varwidth}
\usepackage{tabularx}
\usepackage{siunitx}
\usepackage{algpseudocode,algorithm,algorithmicx}
\usepackage{url}
\usepackage{mathtools}
\usepackage{commath}
\usepackage{pdflscape}
\usepackage{siunitx}
\usepackage{enumitem}

\usepackage{booktabs}

\usepackage{makecell}

\algnewcommand{\IIf}[1]{\State\algorithmicif\ #1\ \algorithmicthen}
\algnewcommand{\EndIIf}{\unskip\ \algorithmicend\ \algorithmicif}

\newcolumntype{R}[2]{%
	>{\adjustbox{angle=#1,lap=\width-(#2)}\bgroup}%
	l%
	<{\egroup}%
}
\usepackage[utf8]{inputenc}
\usepackage[T1]{fontenc}
\usepackage[english]{babel}
\usepackage{lmodern}

\newtheorem{lemma}{Lemma}
\newtheorem{theorem}{Theorem}
\newtheorem{proposition}{Proposition}
\newtheorem{example}{Example}
\newtheorem{corollary}{Corollary}
\newtheorem{property}{Property}

\begin{document}

\begin{frontmatter}

\title{Unsupervised Anomaly Detection in Stream Data with Online Evolving Spiking Neural Networks}

\author[address1]{Piotr S. Maciąg\corref{mycorrespondingauthor}}
\cortext[mycorrespondingauthor]{Corresponding author}
\ead{pmaciag@ii.pw.edu.pl}

\author[address1]{Marzena Kryszkiewicz}
%\cortext[mycorrespondingauthor2]{Corresponding author}
\ead{mkr@ii.pw.edu.pl}

\author[address1]{Robert Bembenik}
%\cortext[mycorrespondingauthor3]{Corresponding author}
\ead{r.bembenik@ii.pw.edu.pl}

\author[address2]{Jesus~L.~Lobo}
\ead{jesus.lopez@tecnalia.com}

\author[address2,address3]{Javier~Del~Ser}
\ead{javier.delser@tecnalia.com}

\address[address1]{Warsaw University of Technology, Institute of Computer Science\\Nowowiejska 15/19, 00-665, Warsaw, Poland}
\address[address2]{TECNALIA, Basque Research and Technology Alliance (BRTA)\\Parque Tecnológico de Bizkaia, E-700, 48160 Derio, Spain}
\address[address3]{University of the Basque Country (UPV/EHU)\\48013 Bilbao, Spain}

% \author{Piotr~S.~Maciąg$^{\textsuperscript{\orcidicon{0000-0001-5486-7927}}}$,
%         Marzena~Kryszkiewicz$^{\textsuperscript{\orcidicon{0000-0003-4736-4031}}}$,
%         Robert~Bembenik$^{\textsuperscript{\orcidicon{0000-0002-7771-2351}}}$,
%         Jesus~L.~Lobo$^{\textsuperscript{\orcidicon{0000-0002-6283-5148}}}$,
% Javier~Del~Ser$^{\textsuperscript{\orcidicon{0000-0002-1260-9775}}}$,~\IEEEmembership{Senior Member,~IEEE}% <-this % stops a space
% %Manuscript received November 28, 2019; revised --.--.--.
% \thanks{P. Maciąg, M. Kryszkiewicz and R. Bembenik are with the Institute of Computer Science, Warsaw University of Technology, Nowowiejska 15/19, 00-665, Warsaw, Poland (e-mail: pmaciag@ii.pw.edu.pl, mkr@ii.pw.edu.pl, r.bembenik@ii.pw.edu.pl).}% <-this % stops a space
% \thanks{J.~L.~Lobo is with TECNALIA. Parque Tecnológico de Bizkaia, c/ Geldo, 48160 Derio, Spain (e-mail: jesus.lopez@tecnalia.com).}
% \thanks{J. Del Ser is with TECNALIA. Parque Tecnológico de Bizkaia, c/ Geldo, 48160 Derio, Spain and
% University of the Basque Country UPV/EHU, 48013 Bilbao, Spain (e-mail: javier.delser@tecnalia.com).} % <-this % stops a space
% \thanks{\textit{(Corresponding author: Piotr S. Maciąg.)}}
% }

%\maketitle

\begin{abstract}
Unsupervised anomaly discovery in stream data is a research topic with many practical applications. However, in many cases, it is not easy to collect enough training data with labeled anomalies for supervised learning of an anomaly detector in order to deploy it later for identification of real anomalies in streaming data. It is thus important to design anomalies detectors that can correctly detect anomalies without access to labeled training data.
Our idea is to adapt the Online evolving Spiking Neural Network (OeSNN) classifier to the anomaly detection task. As a result, we offer an Online evolving Spiking Neural Network for Unsupervised Anomaly Detection algorithm (OeSNN-UAD), which, unlike OeSNN, works in an unsupervised way and does not separate output neurons into disjoint decision classes. OeSNN-UAD uses our proposed new two-step anomaly detection method. Also, we derive new theoretical properties of neuronal model and input layer encoding of OeSNN, which enable more effective and efficient detection of anomalies in our OeSNN-UAD approach. The proposed OeSNN-UAD detector was experimentally compared with state-of-the-art unsupervised and semi-supervised detectors of anomalies in stream data from the Numenta Anomaly Benchmark and Yahoo Anomaly Datasets repositories. Our approach outperforms the other solutions provided in the literature in the case of data streams from the Numenta Anomaly Benchmark repository. Also, in the case of real data files of the Yahoo Anomaly Benchmark repository, OeSNN-UAD outperforms other selected algorithms, whereas in the case of Yahoo Anomaly Benchmark synthetic data files, it provides competitive results to the results recently reported in the literature.
\end{abstract}


\end{frontmatter}

%\linenumbers

\section{Introduction}
\label{sec:Introduction}

Unsupervised anomaly discovery in stream data is a research topic that has important practical applications. For example, an Internet system administrator may be interested in recognition of abnormally high activity on a web page potentially caused by a hacker attack. Unexpected spiking usage of a CPU unit in a computer system could be another example of anomalous behaviour that may require investigation. Correct detection and classification of such anomalies may enable optimization of the performance of the computer system. However, in many cases, it is not easy to collect enough training data with labeled anomalies for supervised learning of an anomaly detector in order to use it later for identification of real anomalies in streaming data. It is thus particularly important to design anomalies detectors that can correctly detect anomalies without access to labeled training data.
Moreover, since the characteristic of an input data stream may be changing, the anomaly detector should learn in an online mode.

In order to design an effective anomaly detection system, one can consider adaptation of evolving Spiking Neural Networks (eSNNs) \citep{Kasabov2014,Lobo108,Lobo2020,Maciag2020-IJCNN2020} to the task. eSNN is a neural network with an evolving repository of output neurons, in which learning processes, neuronal communication and classification of data instances are based solely on transmission of spikes from input neurons to output neurons \citep{Kasabov2014}. The spikes increase so called post-synaptic potential values of output neurons, and directly influence the classification results. The input layer of neurons in eSNN transforms input data instances into spikes. Depending on the type of input data, the transformation can be carried out by means of the temporal encoding methods such as Step-Forward or Threshold-Based \citep{Petro2019_TNNLS_SpikeEncodingMethods,Maciag2019_AirPollutionESNN} or, alternatively, with the use of Gaussian Receptive Fields \citep{Lobo108}. The distinctive feature of the eSNN is that its repository of output neurons evolves during the training phase based on candidate output neurons that are created for every new input data sample \citep{Kasabov2013,Kasabov2015}. More specifically, for each new input value, a new candidate output neuron is created and is either added to the output repository or, based on the provided similarity threshold, is merged with one of the output neurons contained in the repository.

Recently, an online variant OeSNN of eSNN was proposed for classification of stream data \citep{Lobo108}. Contrary to the eSNN architecture, the size of the evolving repository of output neurons in OeSNN is limited. When the repository of output neurons is full and a new candidate output neuron is significantly different from all of the neurons in the repository, an oldest neuron is replaced with the new candidate output neuron. It was claimed in \citep{Lobo108} that OeSNN is able to make fast classification of input stream data, while preserving restrictive memory limits. Considering all the positive features of eSNN and OeSNN, in this article, we offer a novel Online evolving Spiking Neural Network for Unsupervised Anomaly Detection (OeSNN-UAD) in stream data.

Our main contributions presented in this article are as follows:

\begin{itemize}
    \item We offer a new OeSNN-UAD anomaly detector working online in an unsupervised way. It adapts the architecture of OeSNN, which also works in an online way, but, unlike OeSNN-UAD, requires supervised training. The main distinction between our detector and OeSNN lies in applying different models of an output layer and different methods of learning and input values classification. While output neurons of OeSNN are divided into separate decision classes, there is no such separation of output neurons in our approach. Rather than that, each new output neuron is assigned an output value, which is first randomly generated based on recent input values and then is updated in the course of learning of OeSNN-UAD.

    \item As a part of the proposed OeSNN-UAD detector, we offer a new anomaly classification method, which classifies an input value as anomalous only in the following two cases:
    \begin{enumerate}
        \item if none of output neurons in the repository fires, or otherwise,
        \item if an error between an input value and its OeSNN-UAD prediction is greater than the average prediction error plus user-given multiplicity of the standard deviation of the recent prediction errors.
    \end{enumerate}
    This two-step approach to classification of an input value as anomalous or not enables more effective detection of anomalies in input stream data and to the best of our knowledge was not previously used in the literature.

    \item We derive the important theoretical property of the OeSNN neuronal model that shows that the values of post-synaptic potential thresholds of all output neurons are the same. This property is inherited by our OeSNN-UAD detector. The obtained result eliminates the necessity of recalculation of these thresholds when output neurons of OeSNN, as well as of OeSNN-UAD, are updated in the course of the learning process, and increases the speed of classification of input stream data. Moreover, we also prove that firing order values of input neurons do not depend on values of $TS$ and $\beta$ parameters, which were previously used in OeSNNs for input value encoding with Gaussian Receptive Fields.

	\item We prove experimentally that in the case of stream data from the Numenta Anomaly Benchmark repository \citep{NAB2019} as well as from the Yahoo Anomaly Datasets repository \citep{YahooAnomalyDataset} the proposed OeSNN-UAD detects anomalies in unsupervised way more effectively than other state-of-the-art unsupervised and semi-supervised detectors proposed in the literature.

	\item Eventually, we argue that the proposed OeSNN-UAD is able to make fast detection  of anomalies among data stream input values and works efficiently in environments with imposed restrictive memory limits.
\end{itemize}

The paper is structured as follows. In Section~\ref{sec:Related Work}, we overview the related work. In Section~\ref{sec:OeSNN}, we present the architecture of Online evolving Spiking Neural Networks, whose adaptation proposed by us will be then used in OeSNN-UAD.
In Section~\ref{sec:TheorethicalProp}, we provide new theoretical properties of neuronal model and input layer encoding of OeSNN, which enable more effective and efficient detection of anomalies in our OeSNN-UAD approach. In Section~\ref{sec:OeSNN-UAD}, we offer our online method to unsupervised anomaly detection in stream data OeSNN-UAD. Section~\ref{sec:Learning} presents and discusses the proposed OeSNN-UAD algorithm in detail. In Section~\ref{sec:Experimental Results}, we present the results of comparative experimental evaluation of the proposed OeSNN-UAD detector and state-of-the-art unsupervised and semi-supervised detectors of anomalies. We conclude our work in Section~\ref{sec:Conclusions and Future Work}.

%\newpage
\section{Related Work}
\label{sec:Related Work}

%In this section, we first review methods of unsupervised anomaly detection in stream data (with particular emphasis on the methods reported in the experimental evaluation). Then we recall related work in anomaly detection with eSNN.

%Unsupervised anomaly detection in time series data is an important task, which attracts attention of researchers and practicioners.
%The OeSNN-UAD approach presented in our article was experimentally compared against the other state-of-the-arts methods and algorithms using data files of two benchmarks: Numenta Anomaly Benchmark \citep{Ahmad2017} and Yahoo Anomaly Dataset \citep{YahooAnomalyDataset}. The methods or algorithms used for comparison are:%
A number of solutions to the task of unsupervised anomaly detection in time series data was offered in the literature. The state-of-the-art algorithms for unsupervised and semi-supervised anomaly detection are:

\begin{itemize}
    \item Numenta and NumentaTM \citep{Ahmad2017} - two slightly different algorithms that consist of the following modules: (i) a Hierarchical Temporal Memory (HTM) network for predicting the current  value of an input stream data, (ii) an error calculation module, and (iii) an anomaly likelihood calculation module, which classifies the input value as an anomaly or not based on the likelihood of the calculated error. Both algorithms are implemented in Python and are available as a part of the Numenta Anomaly Benchmark repository. NumentaTM and Numenta differ in implementation of the HTM network and its parameters initialization.

    \item HTM JAVA \citep{Hawkins2016} - a JAVA implementation of the Numenta algorithm.

    \item Skyline \citep{Skyline} - an algorithm based on ensembles of several outliers' detectors, such as e.g. Grubb's test for outliers or a simple comparison of the current input value of a data stream against the deviation from the average of past values. In Skyline, a given input value of a data stream is classified as an anomaly if it is marked as anomalous by the majority of ensemble detectors. Skyline is implemented in Python and is available as a part of the Numenta Anomaly Benchmark repository.

    \item TwitterADVec \citep{TwitterAD2015} - a method for anomaly detection based on the Seasonal Hybrid ESD (S-H-ESD) algorithm \citep{Chandola2009}. For given time series values, the S-H-ESD algorithm first calculates extreme Student deviates \citep{Rosner1983} of these values and then, based on a statistical test, decides which of these values should be marked as outliers. The TwitterADVec method is currently implemented as an R language package and is a part of the Numenta Anomaly Benchmark repository.

    \item Yahoo EGADS (Extensible Generic Anomaly Detection System) \citep{Laptev2015} -  an algorithm consisting of the following modules: (i) a time-series modeling module, (ii) an anomaly detection module, and (iii) an alerting module. Yahoo EGADS is able to discover three types of anomalies: outliers, sudden changepoints in values and anomalous subsequences of time series. To this end, the following three different anomaly detectors were implemented in Yahoo EGADS: (i) time series decomposition and prediction for outliers' detection, (ii) a comparison of values of current and past time windows for changepoint detection, and (iii) clustering and decomposition of time series for detection of anomalous subsequences.

    \item DeepAnT \citep{Munir2019_IEEEAccess_DEEPAnT} - a semi-supervised anomaly detection method based on either convolutional neural networks or Long-Short Term Memories networks. DeepAnT consists of a time series prediction module and an anomaly detection module.
    %Contrary to the OeSNN-UAD detector we propose in this article, which detects anomalies in the whole time series,
    DeepAnT uses the first part of a time series as a training and validation data and detects anomalies in the remaining part of the time series.
    % The advantage of our OeSNN-UAD method over DeepAnT is its ability to learn the correct classification of anomalies based on the whole provided time series, rather than only from its training part.

    \item Bayesian Changepoint \citep{Adams2007_CoRR_BayesChangePT} - an online algorithm for sudden changepoint detection in time series data by means of the Bayesian inference. This method is particularly suited to such time series data, in which it is possible to clearly separate partitions of values generated from different data distributions. The algorithm is able to detect the most recent changepoint in the current input values based on the analysis of probability distributions of time series partitions, which are created from changepoints registered in the past values.

    \item EXPected Similarity Estimation (EXPoSE) \citep{Schneider2016} - an algorithm that classifies anomalies based on the deviation of an input observation from an estimated distribution of past input values.

    \item KNN CAD \citep{Burnaev2016} - a method of anomaly detection in univariate time series data based on nearest neighbors classification. KNN CAD method first transforms time series values into its Caterpillar matrix. Such a matrix is created both for the most recent input value (which is classified as an anomaly or not) and for a sequence of past values, which are used as reference data. Next, the Non-Conformity Measure (NCM) is calculated both for the classified value and for the reference values using the created Caterpillar matrix. Eventually, the anomaly score of the classified input value is obtained by comparing its NCM with NCMs of the reference values.

    \item Relative Entropy \citep{Wang2011} - a method, which uses a relative entropy metric (Kullback-Leibler divergence) of two data distributions to decide if a series of input values can be classified as anomalies.

    \item ContextOSE \citep{Contextual} - an algorithm that creates a set of contexts of time series according to the characteristics of its values. A subsequence of most recent input values is classified as anomalous if its context differs significantly from the contexts of past subsequences of values.

\end{itemize}

The above presented methods and algorithms are directly compared to our approach in the experimental evaluation provided in Section~\ref{sec:Experimental Results}. In addition to these approaches, other non-online unsupervised methods of anomaly detection in time series data were proposed. In \citep{Munir2019_Sensors}, an unsupervised detector of anomalies in time series, which combines the ARIMA (Auto-regressive Moving Average) method and convolutional neural networks, was provided. \citep{Ergen2019} introduced an unsupervised anomaly detection method integrating Long-Short Term Memory networks and One-class Support Vector Machines. Yet another non-online unsupervised approach to anomaly detection was offered in \citep{Lin2019-AnomalyDetectionMethod}. It uses a sliding window data stream sampling algorithm based on data elements to sample the sensor network data stream. The sampling result is used as the sample set of the clustering algorithm to detect anomalies in the data stream. %The anomaly detection method was tested on the KDD CUP 99 data stream and was found superior in terms of false negative rate to a sliding window hierarchical clustering algorithm used in the comparison by the authors.%

A supervised eSNN approach to anomaly detection, called HESADM, was proposed in \citep{Demertzis2014_E-Democracy_eSNNAnomaly}. In this approach, the eSNN network is first taught based on a training part of data, and then is used for detection of anomalies in the remaining part of data. In \citep{Demertzis2019}, a semi-supervised approach to anomaly detection with one-class eSNN was offered and dedicated to intrusion detection systems. Contrary to the approaches presented in \citep{Demertzis2014_E-Democracy_eSNNAnomaly,Demertzis2019}, OeSNN-UAD approach offered in this work learns to recognize anomalies in an unsupervised mode, in which anomaly labels are not assigned to data samples.

The distinguishing feature of our proposed OeSNN-UAD anomaly detector compared to the above-mentioned methods and algorithms is that it is the only eSNN-based detector operating both in an online and unsupervised way.

The anomaly detectors proposed in \citep{Zhang2019-AnomalyDetectionSliding} and \citep{Bovenzi2011} are also online unsupervised ones. They detect whether a current data stream value is an anomaly or not based only on a given number of recent input values.

\begin{itemize}
    \item The SPS algorithm, presented in \citep{Bovenzi2011}, uses a window of recent values to calculate statistics that enable dynamic determination of a lower bound and an upper bound on the expected value of the current data point. The real current value is defined as anomalous if it is outside the current bounds. The SPS algorithm was found in \citep{Zoppi2019-SlidingWindowAnomaly} as inferior to non-online unsupervised algorithms kMeans and kHOBS on all datasets tested there (KDD Cup 99 (1999), NSL-KDD (2009), ISCX (2012), UNSW-NB15 (2015)).

    \item The algorithm offered in \citep{Zhang2019-AnomalyDetectionSliding} splits the window of recent values into disjoint subwindows. A vector (PDD) of Probability Density-based Descriptors is calculated for each subwindow. Checking if the current value is an anomaly or not is based on the distances between PDDs of each two consecutive subwindows. The algorithm was tested in \citep{Zhang2019-AnomalyDetectionSliding} on a number of data streams from the Numenta Anomaly Benchmark repository and was found an effective anomaly detector there. In the experimental evaluation provided in Section~\ref{sec:Experimental Results}, we compare its quality with the quality of OeSNN-UAD.
\end{itemize}

An important feature distinguishing OeSNN-UAD from the two above mentioned algorithms is that the predictions made by OeSNN-UAD are not only based on the contents of a sliding window of recent input values, but also on the state of an evolving spiking neural network, which plays a role of an adaptable memory of historical input values.

Overview of anomaly detection techniques for stream data can be found in \citep{Chandola2008,Chandola2009,Chandola2012,Pimentel2014,Chalapathy2019_CoRR_DeepLearningAnomaly,Kwon2017_ClusterComputing_DeepLearningAnomaly,Izakian2013b,Izaian2014,Ergen2019}.

\section{Online Evolving Spiking Neural Networks }
\label{sec:OeSNN}

Our approach to unsupervised anomaly detection adapts the architecture of OeSNN networks previously introduced in \citep{Lobo108}. Thus, the presentation of our approach is preceded with an overview of the architecture of OeSNN network and its classification principles. Input values encoding method and output neuronal model used in that type of network are given as well.

\subsection{Architecture of Online Evolving Spiking Neural Networks}

\begin{figure*}[h]
	\centering
	\includegraphics[width=1.0\linewidth]{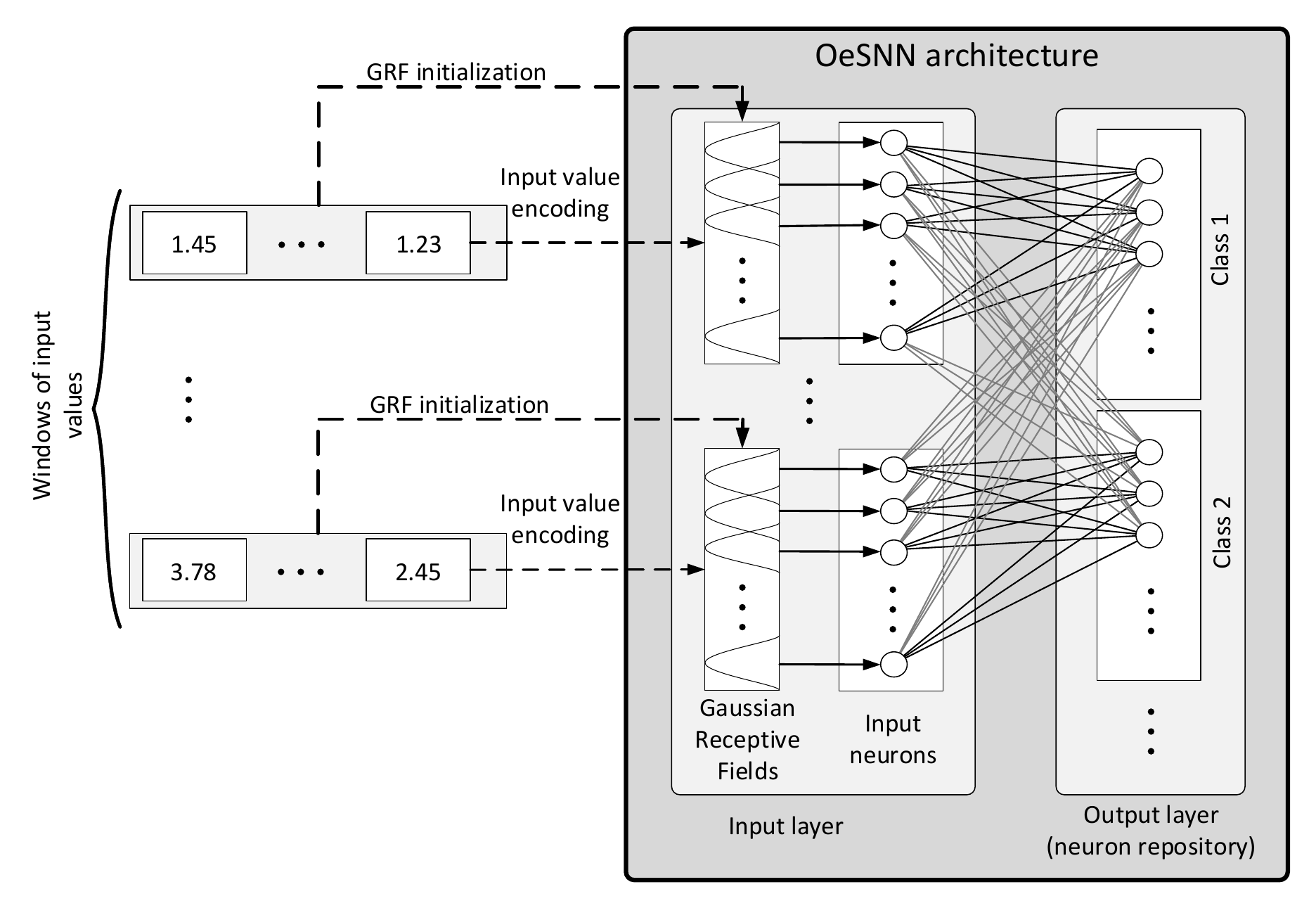}
	\caption{OeSNN architecture introduced in \citep{Lobo108}.}
	\label{Fig:OeSNN-Architecture}
\end{figure*}

OeSNN networks were designed to perform classification tasks in streaming data. OeSNN extends eSNN architecture, which was designed for classification of batch data with separate training and testing parts \citep{KASABOV_07}. Both eSNN and OeSNN architectures consist of input and output layers, however the number of output neurons in OeSNN is limited, while in eSNN it is unlimited. The limitation on the number of neurons in OeSNN is motivated by requirements for the classification of data stream values, where usually a large number of input data is processed and strict memory requirements exist. Both eSNN and OeSNN create a candidate output neuron for each new data sample and either insert it to the output repository when the candidate output neuron is significantly different from all output neurons in the repository or, otherwise, merge it with the most similar output neuron in the repository. However, OeSNN unlike eSNN, controls the size of the repository and when the repository is full, OeSNN inserts the candidate output neuron into it only after the removal of an oldest output neuron from the repository.
The input layer of OeSNN consists of so-called Gaussian Receptive Fields (GRFs) and input neurons. The aim of GRFs is to encode input values into firing times and firing order values of input neurons \citep{Lobo2020-StimuliEncoding}.\footnote{Several published studies have widely agreed on the suitability of the Gaussian Receptive Fields encoding (and its variants) for streaming scenarios (please see, for example, \citep{Lobo2020-StimuliEncoding} or \citep{Petro2019_TNNLS_SpikeEncodingMethods}).
Other alternatives include temporal encoding techniques, such as \textit{Threshold-based Representation algorithm}, \textit{Bens Spiker algorithm} or \textit{Moving Window algorithm}, which were reviewed e.g. in \citep{Petro2019_TNNLS_SpikeEncodingMethods} and in \citep{Maciag2019_AirPollutionESNN}. The temporal encoding techniques can be especially useful in the case of time series data with significant changes of input values, such as EEG or fMRI time series data.
Recently, a new encoding technique was proposed in \citep{Maciag2020-IJCNN2020}. This encoding technique directly calculates firing order values of input neurons (without calculation of exact firing times of input neurons).} The firing order values of input neurons are further used to initialize synapses weights between each input neuron and a candidate output neuron and then to classify an input data sample. The classification in OeSNN is performed by calculation of so-called Post-synaptic Potential (PSP) values of output neurons in the output repository. For each input sample to be classified, first GRFs are initialized, and then they are used for calculation of order values of input neurons. Next, given that encoding the PSP values of output neurons are updated. A decision class assigned to the output neuron whose PSP value first exceeds PSP threshold is returned as a decision class of the input sample. The architecture of OeSNN is presented in Figure~\ref{Fig:OeSNN-Architecture}.
%The distinctive features of OeSNN are fast classification of input samples of data streams and low memory requirements.

OeSNN network enables classification of both univariate as well as multivariate time series data. %In OeSNN, each time series is also referred to as an attribute and has a dedicated group of GRFs and input neurons used to encode input values of that attribute.
The OeSNN classifies each new input sample of data, which consists of only the newest value of each time series. However, the encoding of an input sample is carried out by the input layer using the contents of windows with predefined number of recent values of respective time series.

In the following subsections, we overview the input values encoding technique used in OeSNN as well as its output neuronal model, as the OeSNN-UAD approach proposed in this article  adopts both of these principles in its learning and classification working schema.

\subsection{Input Layer of the OeSNN Network}
\label{subsection:OeSNN-Input Layer}

Since our approach to anomaly detection, which is presented in the following sections, operates on a single time series (that is, a single data stream of values), in this subsection we recall the encoding technique of OeSNN for a single single time series, as it was presented in \citep{Lobo108}. Please note however, that the notions presented here can be easily extended for more than one time series.

The OeSNN network consists solely of an input layer and an output layer. The input layer contains a fixed number of input neurons and their \textit{Gaussian Receptive Fields} (\textit{GRFs}). The output layer, called an \textit{evolving repository}, contains output neurons, whose maximal number is limited. The set of input neurons is denoted by $\mathbf{NI}$, while the set of output neurons by $\mathbf{NO}$. The number of input neurons is determined by user-given parameter $NI_{size}$, whereas the maximal number of output neurons is given by $NO_{size}$, which is also a user-specified parameter value. Each input neuron is linked by a synapse to each output neuron. Let $\mathbf{X}$ denote an input stream of values to be classified and $x_t$ denote the newest value of that stream, which will be subject to classification.

By ${\mathcal{W}}$ we denote a window
%of a fixed number of most recent input values of input data stream $\mathbf{X}$. $\mathcal{W}$ contains
containing the newest value $x_t$ of data stream $\mathbf{X}$, as well as previous $\mathcal{W}_{size} - 1$ values of that data stream. $\mathcal{W}_{size}$ is a user given parameter, which denotes the \textit{size of window} $\mathcal{W}$. Clearly, $\mathcal{W}$ contains $[ x_{t-(\mathcal{W}_{size}-1)}, x_{t-(\mathcal{W}_{size}-2)}, \dots, x_t]$ values of data stream. Two values in the window denoted by $I_{min}^{\mathcal{W}}$ and $I_{max}^{\mathcal{W}}$ play an important role in the classification of the input value $x_t$. $I_{min}^{\mathcal{W}}$ is defined as the minimal value in window ${\mathcal{W}}$, whereas $I_{max}^{\mathcal{W}}$ is defined as the maximal value in ${\mathcal{W}}$. The two values are used to initialize $NI_{size}$ distinct GRFs \textit{excitation} functions - one function per one input neuron. The values of excitation functions obtained for $x_t$ are used to calculate \textit{firing times} and \textit{firing order values} of input neurons, and thus influence the classification result.

%In the GRF's approach, a set of $NI_{size}$ activation functions (e.g. firing time functions or firing order functions) one per each input neuron is defined based on specific values called \textit{centers} in the interval [$I_{min}^{\mathcal{W}}, I_{max}^{\mathcal{W}}]$ that differ from each other by multiple of value called \textit{width}.

The \textit{excitation function} of $j$-\textit{th} GRF, where $j = 0 \dots, NI_{size} - 1$, for input value $x_t$ is denoted by $Exc_{j}^{GRF}(x_{t})$ and is defined as the following Gaussian function:
\begin{equation}
    Exc_{j}^{GRF}(x_{t}) = \exp\biggl({-\frac{1}{2}\biggl(\frac{x_t - \mu_{j}^{GRF}}{\sigma_{j}^{GRF}}\biggl)^2}\biggl),
    \label{Eq:GRFExcitation}
\end{equation}

\noindent where $\mu_{j}^{GRF}$ stands for $j$-\textit{th} GRF's \textit{mean} which is expressed by Eq.~\ref{Eq:GRFCenter}, and ${\sigma_{j}^{GRF}}$ stands for $j$-\textit{th} GRF's \textit{standard deviation} which is expressed by Eq.~\ref{Eq:GRFWidth}:

\begin{equation}
    \mu_{j}^{GRF} = I_{min}^{\mathcal{W}} + \frac{2j - 3}{2}\bigg(\frac{I^{\mathcal{W}}_{max} - I^{\mathcal{W}}_{min}}{NI_{size}-2}\bigg),
    \label{Eq:GRFCenter}
\end{equation}

\begin{equation}
    \sigma_{j}^{GRF} =
    \frac{1}{\beta}\bigg(\frac{I^{\mathcal{W}}_{max} - I^{\mathcal{W}}_{min}}{NI_{size}-2}\bigg),
    \label{Eq:GRFWidth}
    \text{where } \beta \in [1, 2].
\end{equation}

\noindent The parameter $\beta$ that occurs in the equation defining  $\sigma_{j}^{GRF}$ is used to control the degree to which Gaussian Random Fields overlap.
$\mu_{j}^{GRF}$ is also called a \textit{center value of j-th GRF}, while ${\sigma_{j}^{GRF}}$ is also called its \textit{width}.
%One may easily note that the widths of all GRFs are the same.

Please note that the excitation function $Exc_{j}^{GRF}(x_{t})$ takes greatest values for those GRFs whose center values are closest to $x_{t}$. Input neurons associated with such GRFs will have the greatest impact on prediction results. In an approach in which, e.g. for efficiency reasons, only some of input neurons should be used for prediction rather than all, a subset of input neurons related to GRFs with highest excitation values will be fired. A \textit{firing time function} defined in Eq.~\ref{Eq:FiringTime} assigns earlier firing time values to input neurons associated with GRFs having higher excitation values.

The \textit{firing time function} for input neuron $n_j$, where $j = 0 \dots, NI_{size} - 1$, is denoted by $T_{n_j}(x_{t})$, and is defined as follows:

\begin{equation}
    T_{n_j}(x_t) = TS \cdot \big(1 - Exc_{j}^{GRF}(x_t)\big),
    \label{Eq:FiringTime}
\end{equation}

\noindent where \textit{TS} is a user-given basic synchronization time of firings of input neurons in OeSNN and $TS > 0$.

%{Clearly, as follows from the definition of the firing time function, input neurons with greater excitation will fire before input neurons with smaller excitation.}
%

The firing times of input neurons imply their distinct \textit{firing order values}; namely, input neurons with shorter firing times are assigned smaller firing order values, which are integers in $\{0, \dots {NI_{size}} - 1\}$. The firing order value of input neuron $n_j$ is denoted by $order(n_j)$.

%In OeSNN approach, only the \textit{order} in which input neurons fire is taken into account rather than their precise firing times. Input neurons are assigned distinct order values from set $\{0, \dots {NI_{size}} - 1\}$. Input neurons with shorter firing times are assigned smaller order values.

\begin{example}
Given the window of input values and the input layer as shown in Figure~\ref{Fig:Encoding}, let us consider an example of encoding of value $x_t = 0.5$ into excitation values of GRFs and then into firing times and firing order values of input neurons associated with corresponding GRFs. We assume that the size of window $\mathcal{W}$ is $\mathcal{W}_{size} = 14$ and the GRFs parameters $I^{\mathcal{W}}_{min}$ and $I^{\mathcal{W}}_{max}$ are equal to 0.1 and 1.0, respectively. The input layer contains seven neurons each of which is associated with one distinct Gaussian Random Field. The excitation values of GRFs are determined with GRF overlapping parameter $\beta$ = 1.0 and the firing times of input neurons are calculated with synchronization time $TS$ equal to 1.0. The resulting encoding of input value $x_t = 0.5$ is presented beneath:

\begin{itemize}
    \item $Exc^{GRF}_{0}(0.5) = 0.001 \rightarrow $ $T_{n_0}(0.5) = 0.999 \rightarrow$ $order(n_0) = 6$,
    \item $Exc^{GRF}_{1}(0.5) = 0.024 \rightarrow$ $T_{n_1}(0.5) = 0.976 \rightarrow$ $order(n_1) = 5$,
    \item $Exc^{GRF}_{2}(0.5) = 0.227 \rightarrow$ $T_{n_2}(0.5) = 0.773 \rightarrow$ $order(n_2) = 3$,
    \item $Exc^{GRF}_{3}(0.5) = 0.770 \rightarrow$ $T_{n_3}(0.5) = 0.230 \rightarrow$ $order(n_3) = 1$,
    \item $Exc^{GRF}_{4}(0.5) = 0.962 \rightarrow$ $T_{n_4}(0.5) = 0.038 \rightarrow$ $order(n_4) = 0$,
    \item $Exc^{GRF}_{5}(0.5) = 0.442 \rightarrow$ $T_{n_5}(0.5) = 0.558 \rightarrow$ $order(n_5) = 2$,
    \item $Exc^{GRF}_{6}(0.5) = 0.074 \rightarrow$ $T_{n_6}(0.5) = 0.926 \rightarrow$ $order(n_6) = 4$.
\end{itemize}

%The obtained firing order of input neurons : <$n_4$, $n_3$, $n_5$, $n_2$, $n_6$, $n_1$, $n_0$> indicates their importance in determining which output neurons should be fired (please see subsection \ref{subsection:OeSNN-Neuronal Model}) and directly for the classification based on current input value $x_t = 0.5$. Input neuron $n_4$, whose firing order equals 0, is most important, while input neuron $n_0$, whose firing order equals $NI_{size} - 1$ = 6, is least important. In fact, only some first input neurons will be fired in the found order - the ones whose impact will be sufficient to fire an output neuron.

\begin{figure*}[h!t]
	\centering
	\includegraphics[width=1.0\linewidth]{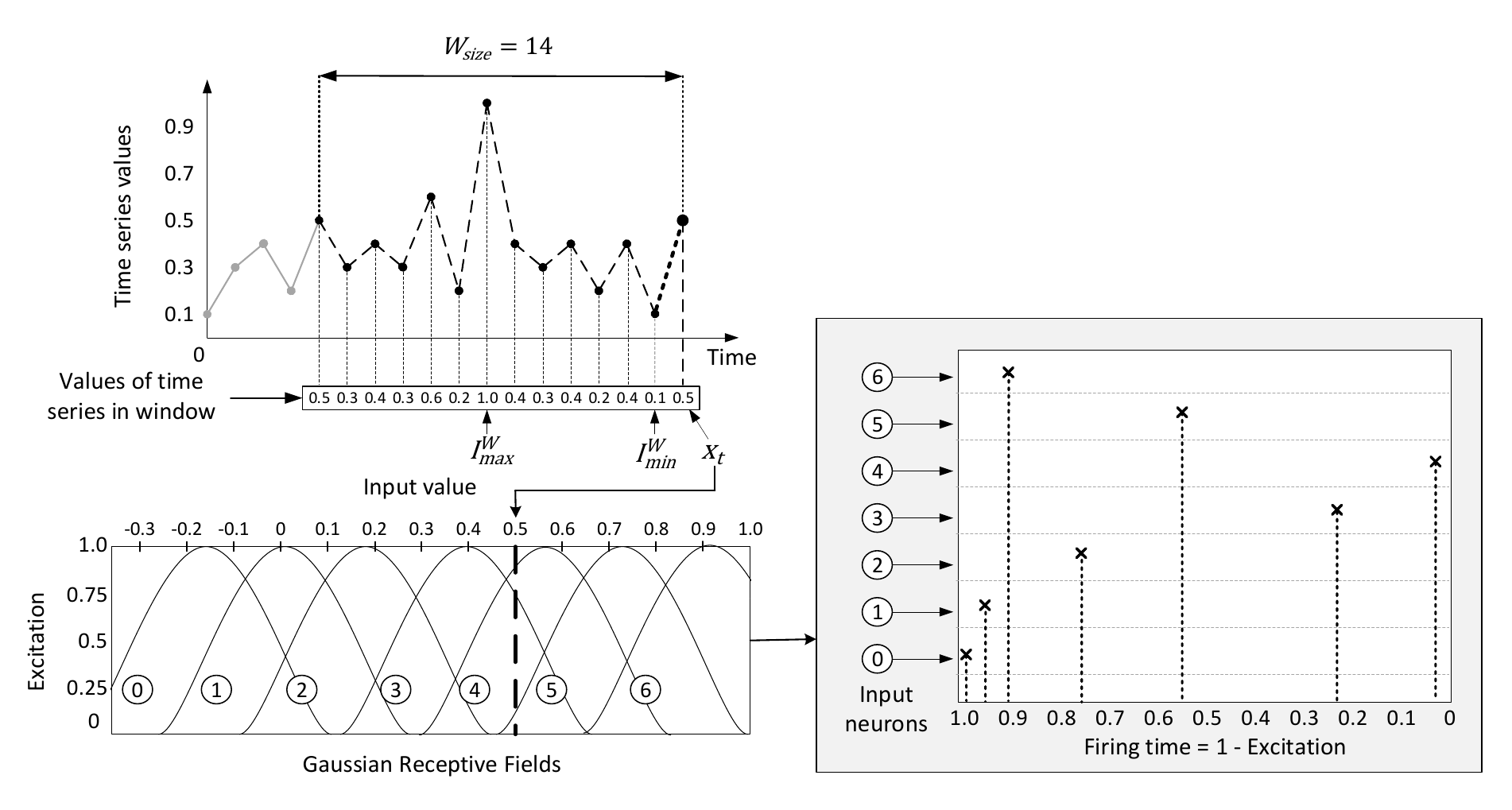}
	\caption{The encoding and the input layer of the proposed network architecture. $\mathcal{W}_{size}$ denotes the size of window $\mathcal{W}$. Current values in $\mathcal{W}$ are used to construct GRFs, while only $x_t$ value is encoded and propagated to neurons in the output repository $\mathbf{NO}$.}
	\label{Fig:Encoding}
\end{figure*}

\label{Example:FiringOrders}
\end{example}

\subsection{Neuronal Model of Output Neurons and Network Learning}
\label{subsection:OeSNN-Neuronal Model}

%In fact, not all input neurons have to be involved in

%{In the proposed approach, the postsynaptic potential of each output neuron $n_i \in \mathbf{NO}$ is reset to 0 (regardless if $n_i$ is actually fired or not) and re-calculated for each input value $x_t \in \mathbf{X}$ being classified. }%

The distinctive feature of OeSNN is the creation of a candidate output neuron for each value $x_t$ of the input data stream. When a new candidate output neuron $n_c$ is created for $x_t$, weights of its synapses are initialized according to input neurons' firing order values obtained as a result of the $x_t$ encoding. The initial weights of synapses between each input neuron $n_j$ in $\mathbf{NI}$ and the candidate output neuron $n_c$ are calculated according to Eq.~\ref{Eq:weights}:

\begin{equation}
    w_{n_jn_c} = mod^{order(n_j)},
    \label{Eq:weights}
\end{equation}
where $mod$ is a user-given modulation factor within range $(0,1)$ and $order(n_j)$ is $n_j$'s firing order value obtained as a result of the $x_t$ encoding.

Vector $[w_{n_0n_c}, \dots,  w_{n_{NI_{size}-1}n_c}]$ of weights of synapses connecting the input neurons in $\mathbf{NI}$ with candidate output neuron $n_c$ will be denoted by $\mathbf{w}_{n_c}$.

%The actual post-synaptic potential threshold $\gamma_{n_c}$ of the candidate output neuron $n_c$ is a fraction of its maximal post-synaptic threshold $PSP^{max}_{n_c}$. The values of both thresholds can be calculated according to Eqs.~(\ref{Eq:PSP_max})~and~(\ref{Eq:PSP_actual}). In Eq.~(\ref{Eq:PSP_actual}), $C$ is a user-given factor within range $(0, 1]$.
%

A candidate output neuron, say ${n_c}$, is characterized also by two additional attributes: the maximal post-synaptic potential $PSP^{max}_{n_c}$ and the post-synaptic potential threshold $\gamma_{n_c}$. The definition of $PSP^{max}_{n_c}$ is given in Eq.~(\ref{Eq:PSP_max}):
\begin{equation}
    PSP_{n_c}^{max} = \sum_{j = 0}^{NI_{size} - 1} w_{n_jn_c}\cdot mod^{order(n_j)},
    \label{Eq:PSP_max}
\end{equation}
where $order(n_j)$ is $n_j$'s firing order value obtained as a result of the $x_t$ encoding.

Property \ref{Property:PSP_nc_max} follows immediately from Eq. \ref{Eq:weights} and Eq. \ref{Eq:PSP_max}.

\begin{property}{}
$PSP_{n_c}^{max} = \sum\limits_{j = 0}^{NI_{size} - 1} mod^{2 \cdot order(n_j)}$.
\label{Property:PSP_nc_max}
\end{property}
%\\

The definition of the post-synaptic potential threshold $\gamma_{n_c}$ is given in Eq.~(\ref{Eq:PSP_actual}):

\begin{equation}
    \gamma_{n_c} = PSP_{n_c}^{max} \cdot C,
    \label{Eq:PSP_actual}
\end{equation}
where \textit{C} is a user fixed value from the interval (0, 1).

\begin{example}
Let us consider again Example \ref{Example:FiringOrders}, which illustrates encoding of input value $x_t = 0.5$  with seven input neurons, as presented in Figure~\ref{Fig:Encoding}. We will show now how synapses weights of a new candidate output neuron are calculated given neuronal model parameters: $mod = 0.5$ and $C = 0.8$. As calculated in Example \ref{Example:FiringOrders}: order(4) = 0, order(3) = 1, order(5) = 2, order(2) = 3, order(6) = 4, order(1) = 5, order(0) = 6. In consequence, the weights of synapses between input neurons and candidate output neuron $n_c$ would be initialized as follows:

\begin{itemize}
    \item $w_{n_4n_c} = 0.5^{0} = 1$,
    \item $w_{n_3n_c} = 0.5^{1} = 0.5$,
    \item $w_{n_5n_c} = 0.5^{2} = 0.25$,
    \item $w_{n_2n_c} = 0.5^{3} = 0.125$,
    \item $w_{n_6n_c} = 0.5^{4} = 0.0625$,
    \item $w_{n_1n_c} = 0.5^{5} = 0.03125$,
    \item $w_{n_0n_c} = 0.5^{6} = 0.015625$.
\end{itemize}

Given these weights of synapses of candidate output neuron $n_c$, the value of its maximal post-synaptic potential $PSP^{max}_{n_c}$  calculated according to Eq.~\ref{Eq:PSP_max} is $1^{2} + 0.5^{2} + 0.25^{2} + 0.125^{2} + 0.0625^{2} + 0.03125^{2} + 0.015625^{2} = 1.333251953$.
\label{Example:SynapticWeights}
\end{example}

In OeSNN, each candidate output neuron, say $n_i$, is either added to the repository $\mathbf{NO}$ or is merged with some output neuron in $\mathbf{NO}$. In fact, each output neuron in $\mathbf{NO}$ is either an extended copy of a candidate output neuron or is an aggregation of a number of candidate output neurons. In comparison with candidate output neurons, output neurons in $\mathbf{NO}$ are characterized by one more attribute - \textit{update counter} $M_{n_i}$, which provides the information from how many candidate output neurons $n_i$ was created. If $M_{n_i} = 1$, then the values of the remaining attributes of $n_i$ are the same as of a former candidate output neuron. Now, each time when an output neuron $n_i$ built from $M_{n_i}$ former candidate output neurons is merged with a current candidate output neuron $n_c$, weight $w_{n_jn_i}$ of the synapse between output neuron $n_i$ and each input neuron $n_j$ is recalculated as shown in Eq.~(\ref{Eq:merged_Weights}), $PSP_{n_i}^{max}$ is recalculated as shown in Eq.~(\ref{Eq:merged_PSP_max}) and $\gamma_{n_i}$ is recalculated according to Eq.~(\ref{Eq:merged_PSP_actual}):
\begin{equation}
    w_{n_jn_i} \gets \frac{w_{n_jn_c}+M_{n_i} \cdot w_{n_jn_i}}{M_{n_i} + 1},
    \label{Eq:merged_Weights}
\end{equation}

\begin{equation}
    PSP_{n_i}^{max} \gets \frac{PSP_{n_c}^{max}+M_{n_i} \cdot PSP_{n_i}^{max}}{M_{n_i} + 1},
    \label{Eq:merged_PSP_max}
\end{equation}

\begin{equation}
    \gamma_{n_i} \gets \frac{\gamma_{n_c}+M_{n_i} \cdot \gamma_{n_i}}{M_{n_i} + 1}.
    \label{Eq:merged_PSP_actual}
\end{equation}

\noindent In addition, $M_{n_i}$ is increased by 1 to reflect the fact that one more candidate output neuron was used to obtain an updated version of output neuron $n_i$.

In the remainder of the article, vector $[w_{n_0n_i}, \dots,  w_{n_{NI_{size}-1}n_i}]$ of weights of synapses connecting the input neurons in $\mathbf{NI}$ with output neuron $n_i$ in $\mathbf{NO}$ will be denoted by $\mathbf{w}_{n_i}$.

In the OeSNN approach, which uses a simplified Leaky Integrate and Fire (LIF) neuronal model of output neurons, output neuron $n_i$ fires for $x_t$ only when its, so called, \textit{post-synaptic potential} is not less than its post-synaptic potential threshold $\gamma_{n_i}$ \citep{Lobo108}.

%OeSNN applies a simplified Leaky Integrate and Fire (LIF) neuronal model of output neurons \citep{Lobo108}. According to LIF, an output neuron accumulates its \textit{Postsynaptic Potential} (PSP) until it reaches an \textit{actual post-synaptic potential} not less than threshold $\gamma$. Then the output neuron fires and its PSP value is reset to 0.
%The accumulation of PSP potential of an output neuron $n_i \in \mathbf{NO}$ is given in Eq.~(\ref{Eq:PSP}):

The \textit{post-synaptic potential} of output neuron $n_i$ in repository $\mathbf{NO}$ for input value $x_t$ is denoted by $PSP_{n_i}$ and is defined by Eq.~(\ref{Eq:PSP}):

\begin{equation}
    PSP_{n_i} = \sum_{j = 0}^{NI_{size} - 1} w_{n_jn_i}\cdot mod^{order(n_j)},
    \label{Eq:PSP}
\end{equation}
where $w_{n_jn_i}$ represents the weight of the synapse linking input neuron $n_j \in \mathbf{NI}$ with output neuron $n_i \in \mathbf{NO}$, and $order(n_j)$ is $n_j$'s firing order value obtained as a result of the $x_t$ encoding.

Classification of an input value, say $x_t$, in OeSNN is performed based on output neurons' post-synaptic potentials   obtained for $x_t$. As presented in Figure~\ref{Fig:OeSNN-Architecture}, output neurons of the output layer are organized into decision classes.
%After the input value is encoded by input layer into firing order of input neurons, actual PSP values of output neurons are calculated according to Eq.~(\ref{Eq:PSP}).
Input value $x_t$ is assigned the decision class of the first output neuron whose post-synaptic potential value exceeded its own post-synaptic potential threshold.

\section{Properties of an OeSNN Neuronal Model}
\label{sec:TheorethicalProp}

In this section, we derive theoretical properties of input and output layers of the OeSNN neuronal model based on definitions provided in Subsection~\ref{subsection:OeSNN-Input Layer} and Subsection~\ref{subsection:OeSNN-Neuronal Model}, respectively:
\begin{itemize}
\item The obtained properties of input layer show that firing order values of input neurons depend neither on values of Gaussian Random Fields overlapping parameter $\beta$ nor on values of basic synchronization time $TS$ of firing of input neurons, and, in consequence, the selection of output neurons to fire does not depend on values of these parameters.
%This finding implies further that looking for optimal values of $\beta$ and \textit{TS} in the case of the OeSNN neuronal model is pointless.
\item The derived properties related to output neurons show that for any recent input value $x_t$, the values of maximal post-synaptic potential of all output neurons are the same and equal to $\sum\limits_{k = 0}^{NI_{size} - 1} mod^{2k}$ = $\frac{1 - mod^{2\cdot NI_{size}}}{1 - mod^2}$ and, in consequence, the values of post-synaptic potential thresholds of all output neurons are the same and equal to $C \cdot \sum\limits_{k = 0}^{NI_{size} - 1} mod^{2k}$ = $C \cdot \frac{1 - mod^{2\cdot NI_{size}}}{1 - mod^2}$. The latter finding enables calculation of the values of the two parameters of output neurons only once; namely, at the beginning of the whole detection anomaly process rather than for each input value of the data stream. As a result, the anomaly detection becomes faster.
\end{itemize}

In fact, the obtained properties are used in our proposed OeSNN-UAD model for efficient detection of anomalies in a data stream (please see the next two sections).

Let us start the presentation of the obtained theoretical results with Proposition \ref{Proposition:Encoding}, which concerns properties of GRFs and input neurons.

%The obtained properties wi ll be used In this subsection, we formulate and prove new properties of the candidate output neurons and the output neurons in $\mathbf{NO}$. The properties presented below, allowed us to simplify the overall design of the algorithms of our OeSNN-UAD approach. The properties presented below extend the notions and properties of output neuronal model presented in Section~\ref{sec:OeSNN} and are used to more efficiently classify input values by our algorithm.https://www.overleaf.com/project/5c4850f94a4fa729f058a322
%

%\begin{proposition}{} For given values $I_{min}^{\mathcal{W}}$, $I_{max}^{\mathcal{W}}$ and $x_t$ of window $\mathcal{W}$ whose most recent value is $x_t$, the following holds for widths and excitation function values of GRFs as well as firing times and firing order values of input neurons calculated for $x_t$:%
%

\begin{proposition}{} For any most recent value $x_t$ of window $\mathcal{W}$ and any input neurons $j_1, j_2 \in \{0, \dots, NI_{size}-1\}$, the following holds:
\begin{enumerate}[label=(\roman*)]
    \item $\sigma_{j_1}^{GRF} = \sigma_{j_2}^{GRF}$ for any values of parameters $\beta$ and \textit{TS}.

    \item $\mu_{j_1}^{GRF}$ does not depend on values of parameters $\beta$ and \textit{TS}.

    \item The truth value of statement "$Exc_{j_1}^{GRF}(x_{t})$ is greater than or equal to $Exc_{j_2}^{GRF}(x_{t})$" does not depend on values of parameters $\beta$ and \textit{TS}.

    \item The following statements are equivalent for any values of parameters $\beta$ and $TS$:
    \begin{itemize}
\item $Exc_{j_1}^{GRF}(x_{t}) \geq Exc_{j_2}^{GRF}(x_{t}).$
\item $T_{n_{j_1}}(x_t) \leq T_{n_{j_2}}(x_t).$
\item $order(n_{j_1}) \leq order(n_{j_2}).$
\end{itemize}

     \item Firing order values of input neurons do not depend on values of parameters $\beta$ and \textit{TS}.
\end{enumerate}

\noindent \textbf{Proof.}
\begin{enumerate}[start=1,label=Ad (\roman*).,leftmargin = 4em]
\item Follows trivially from Eq.~(\ref{Eq:GRFWidth}).

\item Follows trivially from Eq.~(\ref{Eq:GRFCenter}).

\item Follows from  Eq.~(\ref{Eq:GRFExcitation}),  Proposition~\ref{Proposition:Encoding}.(i) and Proposition~\ref{Proposition:Encoding}.(ii).

\item By Eq.~(\ref{Eq:GRFExcitation}), $Exc_{j}^{GRF}(x_{t})$ takes values from interval $[0, 1]$, and by definition of the firing time function for input neuron (see Eq.~(\ref{Eq:FiringTime})), parameter $TS$ may take only a value greater than 0. Hence:
\begin{align} \nonumber
&Exc_{j_1}^{GRF}(x_{t}) \geq Exc_{j_2}^{GRF}(x_{t}) \iff \\ \nonumber
&1 - Exc_{j_1}^{GRF}(x_t) \leq  1 - Exc_{j_2}^{GRF}(x_t) \iff \\\nonumber
&TS \cdot \big(1 - Exc_{j_1}^{GRF}(x_t)\big) \leq  TS \cdot \big(1 - Exc_{j_2}^{GRF}(x_t)\big) \iff \\\nonumber
&T_{n_{j_1}}(x_t) \leq T_{n_{j_2}}(x_t) \iff \\\nonumber
&order(n_{j_1}) \leq order(n_{j_2}). \nonumber
\end{align}
\item Follows immediately from Proposition~\ref{Proposition:Encoding}.(iii) and Proposition~\ref{Proposition:Encoding}.(iv).
$\blacksquare$
\end{enumerate}

\label{Proposition:Encoding}
\end{proposition}

In Lemma~\ref{Lemma:CandidatePSP_max}, we provide properties of candidate output neurons.

\begin{lemma}{}
For each candidate output neuron $n_c$, the following holds:
\begin{enumerate}[label=(\roman*)]

\item Vector $\mathbf{w}_{n_c}$ = $[w_{n_0n_c}, \dots,  w_{n_{NI_{size}-1}n_c}]$ of synapses weights of candidate output neuron  $n_c$ is a permutation of vector $[mod^{0}, mod^{1}, \dots ,$ $mod^{NI_{size} - 1}]$.

\item The sum of all synapses weights of $n_c$ equals $\sum\limits_{k = 0}^{NI_{size} - 1} mod^{k}$.

\item $n_c$'s maximal post-synaptic potential $PSP_{n_c}^{max}$ = $\sum\limits_{k = 0}^{NI_{size} - 1} mod^{2k}$.

\item $n_c$'s post-synaptic potential threshold $\gamma_{n_c}$ = $C \cdot \sum\limits_{k = 0}^{NI_{size} - 1} mod^{2k}$.
\end{enumerate}

\noindent \textbf{Proof.} Let $n_c$ be a candidate output neuron.

\begin{enumerate}[start=1,label=Ad (\roman*).,leftmargin = 4em]
\item By Eq.~(\ref{Eq:weights}), the weight of the synapse linking $n_c$ with input neuron $n_j$ equals $mod^{order(n_j)}$, where $order(n_j)$ is a firing order value of $n_j$. Taking into account that at any time firing order values of input neurons are integers from the set \{0, 1, \dots, $NI_{size}-1$\} and are distinct for each input neuron, vector $\mathbf{w}_{n_c}$ = $[w_{n_0n_c}, \dots,  w_{n_{NI_{size}-1}n_c}]$ of synapses weights of candidate output neuron  $n_c$ is a permutation of vector $[mod^{0}, mod^{1}, \dots ,$ $mod^{NI_{size} - 1}]$.

\item By Lemma~\ref{Lemma:CandidatePSP_max}.(i), the sum of all synapses weights of candidate  output neuron $n_c$  equals $\sum\limits_{k = 0}^{NI_{size} - 1} mod^{k}$.

\item By Property~\ref{Property:PSP_nc_max}, $PSP_{n_c}^{max}$ is the sum of the  squares of all synapses weights of candidate  output neuron $n_c$. Hence, and by Lemma~\ref{Lemma:CandidatePSP_max}.(i), $PSP^{max}_{n_c} $ = $\sum\limits_{k = 0}^{NI_{size} - 1} mod^{2k}$.

\item By Eq.~(\ref{Eq:PSP_actual}) and Lemma~\ref{Lemma:CandidatePSP_max}.(iii), threshold $\gamma_{n_c}$ = $C \cdot \sum\limits_{k = 0}^{NI_{size} - 1} mod^{2k}$. $\blacksquare$
\end{enumerate}
\label{Lemma:CandidatePSP_max}
\end{lemma}

In the remainder of Section~\ref{sec:TheorethicalProp}, we focus on properties of output neurons in repository $\mathbf{NO}$. The derivation of these properties was based on the fact that each output neuron is in fact  constructed from one or more candidate output neurons.

\begin{theorem}{}
Let $n_i$ be an output neuron $n_i$ in repository $\mathbf{NO}$ that was constructed from $M_{n_i}$, where $M_{n_i} \geq 1$, candidate output neurons: $n_{c_1}$, $n_{c_2}$, \dots, $n_{c_{M_{n_i}}}$. The following holds for output neuron $n_i$:

\begin{enumerate}[label=(\roman*)]
\item Vector $\mathbf{w}_{n_i}$ = $[w_{n_0n_i}, \dots,  w_{n_{NI_{size}-1}n_i}]$ of synapses weights of $n_i$ is the average of the vectors of synapses weights of candidate output neurons $n_{c_1}$, $n_{c_2}$, \dots, $n_{c_{M_{n_i}}}$; that is,
\begin{align} \nonumber
\mathbf{w}_{n_i} =
\big[\frac{\sum\limits_{l = 1}^{M_{n_i}} w_{n_0n_{c_l}}}{M_{n_i}} , \dots, \frac{\sum\limits_{l = 1}^{M_{n_i}} w_{n_{NI_{size} - 1}n_{c_l}}}{M_{n_i}}\big]. \nonumber
\end{align}

\item The sum of synaptic weights of $n_i$ equals $\sum\limits_{k = 0}^{NI_{size} - 1} mod^{k}$.

\item $n_i$'s maximal post-synaptic potential $PSP_{n_i}^{max}$ = $\sum\limits_{k = 0}^{NI_{size} - 1} mod^{2k}$.

\item $n_i$'s post-synaptic potential threshold $\gamma_{n_i}$ = $C \cdot \sum\limits_{k = 0}^{NI_{size} - 1} mod^{2k}$.
\end{enumerate}

\noindent \textbf{Proof.}
\begin{enumerate}[start=1,label=Ad (\roman*).,leftmargin = 4em]
\item It follows from  Eq.~(\ref{Eq:merged_Weights}) that the weight $w_{n_jn_i}$ of the synapse linking output neuron $n_i$ with input neuron $n_j$ is the average of the weights of synapses linking candidate output neurons $n_{c_1}$, $n_{c_2}$, \dots, $n_{c_{M_{n_i}}}$ with input neuron $n_j$; that is, $w_{n_jn_i}$ = $\frac{\sum\limits_{l = 1}^{M_{n_i}} w_{n_jn_{c_l}}}{M_{n_i}}$. Hence, $\mathbf{w}_{n_i}$ = $[w_{n_0n_i}, \dots,  w_{n_{NI_{size}-1}n_i}]$ =
$\big[\frac{\sum\limits_{l = 1}^{M_{n_i}} w_{n_0n_{c_l}}}{M_{n_i}} , \dots, \frac{\sum\limits_{l = 1}^{M_{n_i}} w_{n_{NI_{size} - 1}n_{c_l}}}{M_{n_i}}\big].$

\item By Theorem~\ref{Theorem:WeightsSum&Gamma}.(i) and Lemma~\ref{Lemma:CandidatePSP_max}.(ii), the sum of synapses weights of output neuron $n_i$ is equal to $\sum\limits_{k = 0}^{NI_{size} - 1}  w_{n_kn_i}$ =
$\sum\limits_{k = 0}^{NI_{size} - 1} \big( \frac{\sum\limits_{l = 1}^{M_{n_i}} w_{n_kn_{c_l}}}{M_{n_i}} \big)$ =  $\frac{1}{M_{n_i}} \sum\limits_{l = 1}^{M_{n_i}} \big( \sum\limits_{k = 0}^{NI_{size} - 1} w_{n_k}n_{c_l}\big)$ =
 $\frac{1}{M_{n_i}} \sum\limits_{l = 1}^{M_{n_i}} \big( \sum\limits_{k = 0}^{NI_{size} - 1} mod^{k} \big)$ = $ \sum\limits_{k = 0}^{NI_{size} - 1} mod^{k}$.

\item It follows from  Eq.~(\ref{Eq:merged_PSP_max}) that $PSP^{max}$ of output neuron $n_i$ is the average of $PSP^{max}$ of candidate output neurons $n_{c_1}$, $n_{c_2}$, \dots, $n_{c_{M_{n_i}}}$. Hence, and by Lemma~\ref{Lemma:CandidatePSP_max}.(iii), $PSP^{max}_{n_{i}}$ = $\frac{1}{M_{n_i}} \sum\limits_{l = 1}^{M_{n_i}} (PSP^{max}_{n_{i_{l}}})$ = $\sum\limits_{k = 0}^{NI_{size} - 1} mod^{2k}$.

\item It follows from Eq.~(\ref{Eq:merged_PSP_actual}) that  $\gamma_{n_{i}}$ is the average of $\gamma$ thresholds of candidate output neurons $n_{c_1}$, $n_{c_2}$, \dots, $n_{c_{M_{n_i}}}$. Hence, and by Lemma~\ref{Lemma:CandidatePSP_max}.(iv), $\gamma_{n_{i}} = \frac{1}{M_{n_i}} \sum\limits_{l = 1}^{M_{n_i}} (\gamma_{n_{i_{l}}})$ = $\frac{1}{M_{n_i}} \sum\limits_{l = 1}^{M_{n_i}} (C \cdot PSP^{max}_{n_{i_{l}}})$ = $C \cdot \sum\limits_{k = 0}^{NI_{size} - 1} mod^{2k}$.
$\blacksquare $
\end{enumerate}
\label{Theorem:WeightsSum&Gamma}
\end{theorem}

Corollary~\ref{Corollary:WeightsSum&Gamma} follows immediately from Theorem~\ref{Theorem:WeightsSum&Gamma} and the fact that  $\sum\limits_{k = 0}^{NI_{size} - 1} mod^{k}$ and $\sum\limits_{k = 0}^{NI_{size} - 1} mod^{2k}$ are sums of $NI_{size}$ consecutive elements of geometric series.

\begin{corollary}{}
For each output neuron $n_i \in \mathbf{NO}$, the following holds:
\begin{enumerate}[label=(\roman*)]
\item the sum of synaptic weights of $n_i$ equals $\frac{1 - mod^{NI_{size}}}{1 - mod}$,

\item $n_i$'s maximal post-synaptic potential $PSP_{n_i}^{max}$ = $\frac{1 - mod^{2\cdot NI_{size}}}{1 - mod^2}$,

\item $n_i$'s post-synaptic potential threshold $\gamma_{n_i}$ = $C \cdot \frac{1 - mod^{2\cdot NI_{size}}}{1 - mod^2}$.

\end{enumerate}
\label{Corollary:WeightsSum&Gamma}
\end{corollary}

As follows from Lemma~\ref{Lemma:CandidatePSP_max}, Theorem~\ref{Theorem:WeightsSum&Gamma} and Corollary~\ref{Corollary:WeightsSum&Gamma}, all candidate output neurons and output neurons in $\mathbf{NO}$ have the same values of the sum of their synaptic weights, their maximal post-synaptic potentials, and their maximal post-synaptic potential thresholds, respectively. The first two attributes characterizing (candidate) output neurons depend only on the number of input neurons $NI_{size}$ and the value of parameter $mod$, while their third attribute depends also on the value of parameter \textit{C}. The property related to post-synaptic potential thresholds will be used in our proposed algorithm for detecting anomalies.

%\newpage
\section{OeSNN-UAD - the Proposed Anomaly Detection Model Based on Online Evolving Spiking Neural Networks}
\label{sec:OeSNN-UAD}
%In this section, we offer our online OeSNN-UAD approach to unsupervised anomaly detection in stream data, which is based on our adaptation of the OeSNN network. First, we present distinctive features of the proposed OeSNN-UAD approach. Next, we describe in detail its architecture and working principles.

In this section, we offer our online OeSNN-UAD approach to unsupervised anomaly detection in stream data. We strove to design it in such a way so that the following postulates were fulfilled:
\begin{itemize}
    \item Whenever possible, each new input value of a data stream should be correctly classified as either anomalous or non-anomalous.
    \item Each new input value should be used to train OeSNN-UAD for better future classification of input values.
\end{itemize}
 The following subsections present the architecture and working principles of OeSNN-UAD. A step by step presentation of the OeSNN-UAD algorithm is given in Section~\ref{sec:Learning}.

\subsection{The Architecture of OeSNN-UAD}

\begin{figure*}[h!t]
	\centering
	\includegraphics[width=1.0\linewidth]{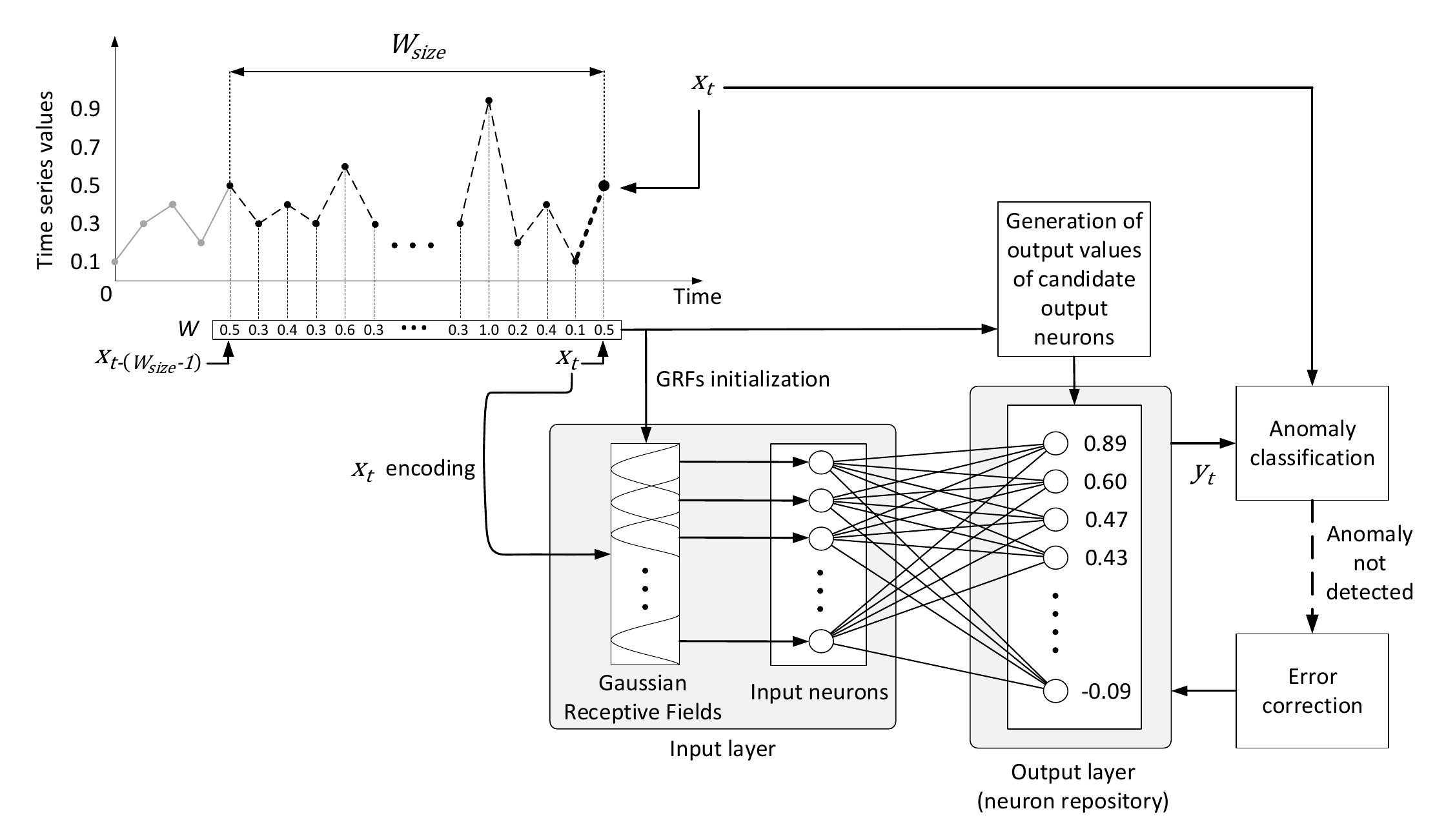}
	\caption{The proposed OeSNN-UAD architecture.
	%The OeSNN-UAD network consists of two layers: input and output.
	}
	\label{Fig:Architecture}
\end{figure*}

The proposed architecture of OeSNN-UAD, which is presented in Figure~\ref{Fig:Architecture}, consists of a modified version of OeSNN and the following three new modules: \textit{Generation of output values of candidate output neurons}, \textit{Anomaly classification} and \textit{Value correction}.

As it can be noted in Figure~\ref{Fig:Architecture}, the output layer of our adapted version of OeSNN network consists of output neurons, which, unlike in OeSNN, are assigned output values, rather than decision classes. In OeSNN-UAD, an output value of each candidate output neuron is first randomly taken from a normal distribution, which is created based on values of the average and standard deviation of input values of window $\mathcal{W}$ and then, in the course of learning of OeSNN-UAD, such candidate is used to update the repository of output neurons. %An initial random value of a candidate output neuron is taken from a normal distribution, which is created based on values of the average and standard deviation of input values of window $\mathcal{W}$.
The output value of an output neuron may be modified in the course of learning of the network in up to two possible ways:
\begin{itemize}
    \item it may be corrected in order to be better adjusted to a current input value of the data stream,
    \item it may be updated with the output value of a current candidate output neuron.
\end{itemize}
The main idea behind this approach is to store the output neurons whose output values correspond to previous non-anomalous  values in the output repository $\mathbf{NO}$ of OeSNN-UAD. In consequence, when non-anomalous input value $x_t$ occurs in a data stream, network prediction value $y_t$ is expected to be similar to value $x_t$, while when anomalous input value $x_t$ occurs, network prediction value $y_t$ is expected to be significantly different from $x_t$.

The OeSNN-UAD anomaly detector works in two phases: in the anomaly detection phase and in the learning phase, which are performed for each input value $x_{t}$ of the data stream:

\begin{enumerate}

\item In the anomaly detection phase, window $\mathcal{W}$ is updated with value $x_t$ and GRFs of input neurons are initialized. The value $x_t$ of $\mathcal{W}$ is used to calculate firing times and firing orders of input neurons in $\mathbf{NI}$. Next, the input value $x_{t}$ is classified as anomalous or not in the following steps. First, output neuron $n_f \in \mathbf{NO}$ that fired as first is obtained. If none of output neurons fired, then input value $x_t$ is immediately classified as an anomaly. Otherwise, the output value of the output neuron that fired as first is reported as network prediction value $y_{t}$ for input value $x_t$. Finally, the \textit{Anomaly classification} module classifies input value $x_t$ as anomalous or not using a prediction error being the absolute difference between $x_t$ and $y_t$ and a threshold value calculated based on errors of recent $\mathcal{W}_{size}$ prediction values.

\item In the network learning phase, new candidate output neuron $n_c$ corresponding to value $x_t$ is created and initialized. The initialization procedure of $n_c$ is performed in three steps: first, the synapses linking $n_c$ with all input neurons in $\mathbf{NI}$ are created and their weights are calculated according to Eq.~\ref{Eq:weights}. Next, \textit{update time} $\tau_{n_c}$ of the candidate output neuron is set to value $t$. Finally, the \textit{Generation of values of candidate output neurons} module assigns initial output value $v_{n_c}$ to the candidate output neuron and update counter $M_{n_c}$ is set to 1. Additionally, if $x_t$ is not classified as anomalous, then the generated initial output value $v_{n_c}$ of the candidate output neuron is corrected by the \textit{Value correction} module.

After  initialization of $n_c$, it is used to update repository $\mathbf{NO}$ of OeSNN-UAD in a similar (but not identical) way as OeSNN repository is updated. $n_c$ is merged with the most similar output neuron $n_s$ already present in $\mathbf{NO}$ for which the Euclidean distance between vectors of synapses weights $\mathbf{w}_{n_c}$ and $\mathbf{w}_{n_s}$ is minimal and less than or equal to the user given threshold $sim$, provided such an output neuron exists. If so, the vector of synapses weights, output value, update time and update counter of $n_s$ are modified according to the formulae given in Eq.~(\ref{Eq:UpdateNeuron}):
\begin{align}
    & \mathbf{w}_{n_s} \gets (\mathbf{w}_{n_s}\cdot M_{n_s} + \mathbf{w}_{n_c})/(M_{n_s} + 1), \nonumber\\
    & v_{n_s} \gets (v_{n_s}\cdot M_{n_s} + v_{n_c})/(M_{n_s} + 1), \nonumber\\
    & \tau_{n_s} \gets (\tau_{n_s}\cdot M_{n_s} + \tau_{n_c})/(M_{n_s} + 1), \nonumber \\
    & M_{n_s} \gets M_{n_s} + 1.
    \label{Eq:UpdateNeuron}
\end{align}
Otherwise, if
%there is no such $n_s$ output neuron in $\mathbf{NO}$,
current size of $\mathbf{NO}$ is less than $NO_{size}$, then $n_c$ is simply added to $\mathbf{NO}$. However, if $\mathbf{NO}$ is full, then $n_c$ replaces the output neuron in $\mathbf{NO}$ whose update time is minimal.
\end{enumerate}

More detailed description of the modules: \textit{Generation of output values of candidate output neurons}, \textit{Anomaly classification} and \textit{Value correction} specific to OeSNN-UAD is provided in the following subsections.

\subsection{Anomaly Classification}
\label{subsubsec:AC}

Given input value $x_{t}$ of the data stream and its prediction $y_{t}$ made by OeSNN-UAD, the aim of the \textit{Anomaly classification} module (see  Figure~\ref{Fig:Architecture}) is to decide whether $x_{t}$ should be classified as an anomaly or not. The approaches proposed in the literature, such as those presented in \citep{Malhotra_ESANN_LSTMAnomalyDetection,Munir2019_IEEEAccess_DEEPAnT,Munir2019_Sensors}, simply calculate an error between the predicted and the real value, compare it against a fixed threshold value and decide if an anomaly occurred. Alternatively, a window of recent predictions errors is used to construct a statistical distribution and obtain the probability of actual prediction error for value $x_{t}$ \citep{Ahmad2017}. If the probability of the actual prediction error is low, the observation is classified as an anomaly. \citep{Carrera2019_PatternRecognition_OnlineAnomaly} takes a different approach and proposes to adapt an error threshold for anomaly classification according to the changing characteristic of the stochastic process generating input data.

In our approach, a vector of error values calculated between predicted values (network responses) and input values of window $\mathcal{W}$ is used to decide if observation $x_{t}$ should be classified as anomalous or not. The error $e_t$ between $x_t$ and its prediction $y_t$ is calculated as the absolute difference between these two values: $e_t = \abs{x_t - y_t}$. Let $\mathbf{e}$ be a subset of set $\{e_{t - (\mathcal{W}_{size} -1)}, \dots, e_{t - 1}\}$ of those $\mathcal{W}_{size}$ prediction error values that were obtained for input values classified as non-anomalous. If $\mathbf{e}$ is not empty, then the mean $\overline{x}_{\mathbf{e}}$ and the standard deviation $s_{\mathbf{e}}$ of error values in $\mathbf{e}$ are calculated and used to classify $x_{t}$ as either an anomaly or not. If the difference between $e_{t}$ and $\overline{x}_{\mathbf{e}}$ is greater than $\varepsilon \cdot s_{\mathbf{e}}$, where $\varepsilon$ is a user-specified anomaly classification factor, then $x_{t}$ is classified as an anomaly, otherwise it is not.

%Given vector $\mathbf{e}$ = $[e_{W_{size}-1}, \dots, e_{t}]$ of prediction error values obtained for recent $\mathcal{W}_{size}$ input values of $\mathbf{X}$ classified so far and vector , their predictions in $\mathbf{Y}$a vector $\mathbf{e}$ of such past $\mathcal{W}_{size}$ error values of $\mathbf{E}$, whose respective input values $x$ were not classified as anomalies is obtained.

If $\mathbf{e}$ is empty, then $x_t$ is classified as non-anomalous. This classification can be justified as follows. Empty set $\mathbf{e}$ indicates that all previous $\mathcal{W}_{size}$ input values were classified as anomalies. This case may indicate the presence of a long anomaly window or the occurrence of a new trend in input values of the data stream. After classifying all values of previous $\mathcal{W}$ window as anomalies, the detector should no longer assume that input values are anomalies, but instead should treat them as normal values and should not indicate the presence of anomalies.

The crucial step of the above procedure is the selection of the value of parameter $\varepsilon$. Usually, the value of that parameter should be experimentally adjusted to the domain of the input data stream. In our experimental evaluation, we performed tuning of values of that parameter.

\subsection{Generation of an Initial Output Value of a Candidate Output Neuron}
\label{subsub:GenerationOutVal}

The module called \textit{Generation of initial output value of a candidate output neuron} in Figure~\ref{Fig:Architecture} is responsible for generating initial output values of candidate output neurons when they are created. The proposed method of generating output values is based on the assumption that network prediction $y_t$ for non-anomalous input value $x_t$ should be relatively similar to $x_t$, whereas network prediction $y_t$ for anomalous $x_t$ should differ from $x_t$ significantly. Following this assumption, we propose to generate initial output values of candidate output neurons based on recent window $\mathcal{W}$ of input values. More specifically, in our approach, the initial output value of the candidate output neuron created for $x_t$ is taken randomly from a normal distribution whose mean and standard deviation are determined based on input values present in $\mathcal{W}$. The aim of this approach is to generate values, which will correspond to the current fluctuations in the data stream.

%Since anomaly classification in OeSNN-UAD is based on prediction $y_t$ of the neural network for input value $x_t$, an initial output value of a candidate output neuron is randomly generated based on recent window $\mathcal{W}$ of input values. The aim of the initial random generation of output values of candidates is to assign to the output neurons values that correspond to the current fluctuations in the data stream. For a non-anomalous input value $x_t$, network prediction $y_t$ should be similar to $x_t$. On the other hand, for an anomalous value $x_t$
%When an anomaly input value $x_t$ (that is a value, which is abnormal with regard to previously observed values) will be classified, then the network response $y_t$ will differ significantly from $x_t$, which will indicate the presence of an anomaly.
%In particular, the initial random output values are in our approach are taken from a normal distribution, which is created from the mean and the standard deviation of values present in $\mathcal{W}$.

\subsection{Correction of an Output Value of an Output Neuron}
\label{subsub:ValueCorr}

The \textit{Value correction} module (see Figure~\ref{Fig:Architecture}) is responsible for correcting the output value of the candidate output neuron $n_c$ created for input value $x_t$. The value correction is performed only when that input value was classified as non-anomalous. Intuitively, one can perceive $n_c$ as a neuron representation of $x_t$, since initial synapses weights of $n_c$ were determined based on firing order values of input neurons for value $x_t$. When $x_t$ is not classified as an anomaly, the initial output value of the candidate neuron $n_c$ is adjusted as follows: $v_{n_c} \leftarrow v_{n_c} + (x_t - v_{n_c})\cdot \xi$. In this formula, $\xi$ is a user given value correction factor within the range $[0, 1]$. If $\xi = 0$, the initial output value of $n_c$ will not change. If $\xi = 1$, then $v_{n_c}$ value will be equal to $x_t$.

Let us assume that output neuron $n_c$ with its output value corrected as described above became output neuron $n_i$ in $\mathbf{NO}$ or was merged with output neuron $n_i$ in $\mathbf{NO}$ and that new input value $x_{t_1}$ arrives at time $t_1$ slightly later than time $t$ and becomes subject to classification. If $x_{t_1}$ is similar to previously classified $x_t$ value, then output neuron $n_i$ corresponding to $x_t$ is very likely to fire as first and its output value $v_{n_i}$ will be reported as network prediction $y_{t_1}$. Since that $v_{n_i}$ output value was previously adjusted to non-anomalous $x_t$ value, then also $x_{t_1}$ value will be likely classified as non-anomalous provided prediction error $e_{t_1}$ is relatively small in comparison with the errors present in $\mathbf{e}$.

Overall, the aim of correcting an output value of candidate output neuron $n_c$ is to prevent future incorrect classification of input values that could be caused by
%random generation of $n_c$'s initial output value based on
possible fluctuations of input values in a data stream.

%Let us assume that new input value $x_{t_1}$ will arrive at time $t_1 > t$ and becomes subject to classification. If $x_{t_1}$ is similar to previously classified $x_t$ value, then the output neuron corresponding to $x_t$ (that is, the output neuron which is either an extension of candidate output neuron $n_c$ or the one which was merged with $n_c$) will fire as first and its output value will be reported as network prediction $y_{t_1}$. Since that output value (and thus network prediction $y_{t_1}$) was previously adjusted to non-anomalous $x_t$ value , then also $x_{t_1}$ would be likely classified as non-anomalous. Overall the aim of this operation is to prevent incorrect classification of anomalous values due to the initial random values of candidate output neurons, while assuming possible fluctuations of input values of a data stream.

%\subsection{Used Notation}

%In Table~\ref{Table:Notation}, we list the notation of parameters used in the formulae and algorithms presented in the article.

\begin{table}[h!t]
\caption{Notations and parameters used in OeSNN-UAD}
\footnotesize{
\resizebox{\columnwidth}{!}{\begin{tabularx}{\linewidth}{lll}
	\toprule
    Notation & Description & Value \\
    \midrule
    $\mathbf{X}$ & Stream of input data & \\
    %$T$ & Number of input values in input stream $\mathbf{X}$ & \\%
    $ \mathcal{W} $ & Window of recent input values & \\
    $ \mathcal{W}_{size} $&  Window size (the number of recent input values in $\mathcal{W}$)  &\\
    $ x_t$ & Input value at time $t$  & \\
    $ y_{t} $ &  OeSNN-UAD prediction of $x_t$   &\\
    $ \mathbf{Y} $ & Vector of predicted values  &\\
    $u_{t}$ & \makecell[l]{Boolean value indicating anomaly \\presence or absence for input value $x_{t}$} & \\
    $\mathbf{U}$ & \makecell[l]{Vector of results of anomaly detection for input values}  &\\
    \midrule
    $ \mathbf{NI} $ & Set of input neurons &\\
    $ NI_{size} $ & Number of input neurons  &\\
    $ TS $ & \makecell[l]{Synchronization time of input neurons firings}  &\\
   % $ \beta $ & Control parameter for GRFs width & [1,2]\\
    $n_{j}$ & $j$-th neuron in the set $\mathbf{NI}$ of input neurons   &\\
    $\mu^{GRF}_{j}$ & \makecell[l]{GRF center for input neuron $n_j$}  &\\
    $\sigma^{GRF}_{j}$ & \makecell[l]{GRF width for input neuron $n_j$}  &\\
    $I_{max}^{\mathcal{W}}$ &\makecell[l]{Maximal input value in window $\mathcal{W}$}  &\\
    $I_{min}^{\mathcal{W}}$ & \makecell[l]{Minimal input value in window $\mathcal{W}$}  &\\
    $Exc_{j}^{GRF}(x_t)$ & Excitation of j-th GRF for value $x_t$  &\\
    $T_{n_j}(x_t)$ & Firing time of input neuron $n_j$ for value $x_t$  &\\
    $\overline{x}_{\mathcal{W}}$, $s_{\mathcal{W}}$ & Mean and standard deviation of input values in $\mathcal{W}$  &\\
    $\mathcal{N}$ & Normal distribution  &\\
    \midrule
    $ \mathbf{NO} $ &  Repository of output neurons  &\\
    $ NO_{size} $ & Number of output neurons in repository $\mathbf{NO}$ &\\
    $ mod $ & \makecell[l]{Modulation factor of weights of synapses} & (0, 1)\\
    $ sim $ & User-given similarity threshold  & %(0, 1]
    \\
    $n_i$ & $i$-th output neuron  from repository $\mathbf{NO}$  &\\
    $\mathbf{w}_{n_i}$ & Vector of synaptic weights of output neuron $n_i$  &\\
    $w_{n_j, n_i}$ & \makecell[l]{Weight of a synapse between $n_{j} \in \mathbf{NI}$ and $n_{i} \in \mathbf{NO}$} &\\
    $\gamma$ & \makecell[l]{Post-synaptic potential threshold of output neurons}  &\\
    $v_{n_i}$ & Output value of output neuron $n_i$  &\\
    $\tau_{n_i}$ & \makecell[l]{Update time of output neuron $n_i$} &\\
    $M_{n_i}$ & Number of updates of output neuron $n_k$  &\\
    $PSP^{max}$ & \makecell[l]{Maximal post-synaptic potential of output neurons}  &\\
    $ C $ & \makecell[l]{Fraction of $PSP^{max}_{n_i}$ for calculation of $\gamma_{n_i}$} & (0, 1]\\
    $n_c$ & New candidate output neuron &\\
    $D_{n_c,n_i}$ & \makecell[l]{Euclidean distance between weights vectors $\mathbf{w}_{n_c}$ and $\mathbf{w}_{n_i}$}  &\\
    $\xi$ & Error correction factor & [0, 1]\\
    \midrule
    $e_t$ & \makecell[l]{Error between input value $x_t$ and its prediction $y_t$}  & \\
    $\mathbf{E}$ & Vector of error values between $\mathbf{X}$ and $\mathbf{Y}$  & \\
    $\varepsilon$ & Anomaly classification factor & $ \geq$ 2 \\
	\bottomrule
\end{tabularx}}
}
\label{Table:Notation}
\end{table}

%\newpage
\section{The OeSNN-UAD Algorithm for Unsupervised Anomaly Detection}
\label{sec:Learning}

In this section, our proposed OeSNN-UAD algorithm (please refer to Table~\ref{Table:Notation} for notation used in it), working principles of which were described in Section~\ref{sec:OeSNN-UAD}, is  presented and discussed in detail. The main procedure of OeSNN-UAD is presented in Algorithm~\ref{Alg:eSNN1}. All OeSNN-UAD input parameters, that is $\mathcal{W}_{size}$, $NI_{size}$, $NO_{size}$, $mod$, $C$, $sim$, $\xi$, $\varepsilon$ are constant during the whole process of anomaly detection and learning.
%(Please note that the values of $\beta$ and $TS$ parameters of GRFs do not affect firing order of input neurons as we proved in Section~\ref{sec:TheorethicalProp} and thus they are not used in the procedure of Algorithm~\ref{Alg:eSNN1}.)
First, the current counter $CNO_{size}$ of output neurons in output repository $\mathbf{NO}$ is set to $0$. Next, based on the fact that the values of post-synaptic potential thresholds $\gamma_{n_i}$ are the same for all output neurons $n_i$ in $\mathbf{NO}$ and depend only on constants $NI_{size}$, $mod$ and $C$ (as follows from Theorem~\ref{Theorem:WeightsSum&Gamma}.(iv) and Corollary~\ref{Corollary:WeightsSum&Gamma}.(iii) provided in Section~\ref{sec:TheorethicalProp}), their common post-synaptic potential threshold, denoted by $\gamma$, is calculated only once. Then, window $\mathcal{W}$ is initialized with input values $x_1, \dots, x_{\mathcal{W}_{size}}$ from data stream $\mathbf{X}$. These values are not classified as anomalies (the assumption that the first $\mathcal{W}_{size}$ values of the data stream are not treated as anomalies is similar to the approach taken in the Numenta Anomaly Benchmark repository, which we use in our experimental evaluation).

The detection of anomalies among input values $x_t$ of $\mathbf{X}$, where $t \geq {\mathcal{W}_{size}+1}$, starts in step~\ref{step:foreachX} of Algorithm~\ref{Alg:eSNN1} and for each of these input values is carried out as follows. First, window $\mathcal{W}$ is updated with input value $x_t$ which becomes subject to anomaly classification, and GRFs as well as firing order values of input neurons are determined based on the content of window $\mathcal{W}$, as presented in Algorithm~\ref{Alg:InitializeGRF}. Next, output neuron $n_f \in \mathbf{NO}$ that fires as first is obtained (see Algorithm~\ref{Alg:FiresFirst}). %and its output value $v_{n_f}$ is reported as the prediction $y_{t}$ of input value $x_{t}$.

\begin{algorithm}[h!t]
	\caption{OeSNN-UAD}
	\small{
	\begin{algorithmic}[1]
	    \Require $\mathbf{X} = [x_1, x_2, \dots, x_{T}]$ - stream of input data.
	    \Statex \textbf{Assure constant:}
	    \Statex $\mathcal{W}_{size}$, $NO_{size}$, $NI_{size}, mod$, $C$, $sim$, $\xi$, $\varepsilon$
		\Ensure $\mathbf{U}$ - a vector with classification of each $x \in \mathbf{X}$ as an anomaly or not.
		\State $CNO_{size} \gets 0$
		%\State $D^{max} \gets \sqrt{\sum\limits_{k = 0}^{NI_{size} - 1}\big(mod^{NI_{size} - 1 - k} - mod^{k}\big)^2}$
		%\State $\gamma \gets C \cdot \sum\limits_{k = 0}^{NI_{size} - 1} mod^{2\cdot k}$
		%
		\State $\gamma \gets C \cdot \frac{1 - mod^{2\cdot NI_{size}}}{1 - mod^2}$
	    \State Initialize $\mathcal{W}$ with $x_{1}, \dots, x_{\mathcal{W}_{size}} \in \mathbf{X}$
	    \State Initialize $y_1, \dots, y_{\mathcal{W}_{size}}$ with random values from $ \mathcal{N}(\overline{x}_{\mathcal{W}}^{ }, s_{\mathcal{W}}^2)$ and add to $\mathbf{Y}$
	    \State Intialize $\mathbf{E}$ with $e_l \gets \abs{x_l - y_l},$ $l = 1, \dots, \mathcal{W}_{size}$
	    \State Set $u_1, \dots, u_{\mathcal{W}_{size}} $  to $False$ and add to $\mathbf{U}$
	    \For {$t \gets \mathcal{W}_{size} + 1$ to $T$} \label{step:foreachX}

	        \Statex ~~~~
	        \Statex ~~~~\{*--------------------- OeSNN-UAD anomaly detection ---------------------*\}
	        \State Update window $\mathcal{W}$ with value $x_{t}$
	        \State \Call{InitializeGRFs}{$\mathcal{W}$}
		    %\State Propagate spikes from $\mathbf{NI}$ to $\mathbf{NO}$
		    \State $n_f \gets$ \Call{FiresFirst}{$CNO_{size}$}
		    \If{$n_f$ is NULL} \Comment{If none of output neurons fired}
		        \State $y_{t} \gets $ NULL; append $y_{t}$ to $\mathbf{Y}$
		        \State $e_{t} \gets +\infty$; append $e_{t}$ to $\mathbf{E}$
		        \State $u_t \gets True$ \Comment{Immediately classify $x_t$ as anomaly}
		    \Else
		    \State $y_{t} \gets v_{n_f}$; append $y_{t}$ to $\mathbf{Y}$
		    \State $e_{t} \gets \abs{x_{t} - y_{t}}$; append $e_{t}$ to $\mathbf{E}$
		    \State $u_{t} \gets$ \Call{ClassifyAnomaly}{$\mathbf{E}, \mathbf{U}$}
		    \EndIf
		    \State Append $u_{t}$ to $\mathbf{U}$

		    \Statex ~~~~
		    \Statex ~~~~\{*------------------------------ OeSNN-UAD learning --------------------------*\}
		    \State Create a candidate output neuron $n_c$
		    \State $n_c$ $\gets$ \Call{InitializeNeuron}{$\mathcal{W}, t$} \label{step:NeuronInit}
		    \If{$u_t = False$} \Comment{Anomaly for $x_t$ not detected}
		        \State $v_{n_c} \gets v_{n_c} + (x_{t} - v_{n_c}) \cdot \xi$ \Comment{Correct generated output value of $n_c$}
		    \EndIf
		    \State $n_s \gets$ \Call{FindMostSimilar}{$n_c$} \label{step:MinDistFor}
		    \If {$D_{n_c, n_s} \leq sim$}
		       \State \Call{UpdateNeuron}{$n_s, n_c$}
		    \ElsIf {$CNO_{size} < NO_{size}$}
		        \State Insert $n_c$ to $\mathbf{NO}$; $CNO_{size} \gets CNO_{size} + 1$
		    \Else
		        \State $n_{oldest} \gets$ an output neuron in $\mathbf{NO}$ such that
		        \par  \hskip\algorithmicindent \hskip\algorithmicindent
		          $\tau_{n_{oldest}} = min\{\tau_{n_i}~|~i = 0, \dots, NO_{size}-1\}$
		        \State Replace $n_{oldest}$ with $n_c$ in $\mathbf{NO}$
		    \EndIf
		\EndFor
		\State \Return $\mathbf{U}$
	\end{algorithmic}
	}
	\label{Alg:eSNN1}
\end{algorithm}

\begin{algorithm}[h!t]
	\caption{\protect \Call{InitializeGRFs}{$\mathcal{W}$}}
	\begin{algorithmic}[1]
	    \Require $\mathcal{W} = \{ x_{t-(\mathcal{W}_{size} - 1)}, \dots, x_t\}$ window of input values of $\mathbf{X}$.
	%	\Procedure{InitializeGRFs}{$\mathcal{W}_t$}
		\State Obtain current $I_{min}^{\mathcal{W}}$ and $I_{max}^{\mathcal{W}}$ from $\mathcal{W}$
		\State Calculate $\sigma^{GRF} \gets \dfrac{I^{\mathcal{W}}_{max} - I^{\mathcal{W}}_{min}}{NI_{size}-2}$
	        \For{$ j \gets 0$ to $NI_{size}-1$} \Comment{For all input neurons in $\mathbf{NI}$}
	            \State $ \sigma_{j}^{GRF} \gets \sigma^{GRF}$  \Comment{By Eq.~(\ref{Eq:GRFWidth}) and Propos.~\ref{Proposition:Encoding}.(i)}
		        \State Calculate $\mu_{j}^{GRF}$ \Comment{By Eq.~(\ref{Eq:GRFCenter})}
		        \State Calculate excitation $Exc_{j}^{GRF}(x_t)$ \Comment{By Eq.~(\ref{Eq:GRFExcitation})}
		        \State Calculate firing time $T_{n_j}(x_t)$ \Comment{By Eq.~(\ref{Eq:FiringTime})}
		    \EndFor
		    \For{$ j \gets 0$ to $NI_{size}-1$}
	            \State Calculate $order(j)$
	        \EndFor
	%	\EndProcedure
	\State \Return
	\end{algorithmic}
	\label{Alg:InitializeGRF}
\end{algorithm}

The determination of the first output neuron $n_f$ to fire is carried out according to our proposed procedure $\Call{FiresFirst}{}$, which is presented in Algorithm~\ref{Alg:FiresFirst}. To this end, for efficiency reasons, the algorithm uses lower approximation $\underline{PSP}_{n_i}$ of post-synaptic potential $PSP_{n_i}$ for each output neuron $n_i \in \mathbf{NO}$ instead of ${PSP}_{n_i}$. Lower approximation $\underline{PSP}_{n_i}$ differs from $PSP_{n_i}$ in that  $PSP_{n_i}$ is obtained after firing all input neurons, while $\underline{PSP}_{n_i}$ sufficiently approximates $PSP_{n_i}$ after firing only a few most significant input neurons, whose firing order values are lowest. %Please note that the first output neuron, say $n_f$, whose $\underline{PSP}_{n_f}$ value exceeds $\gamma$ threshold is not necessarily the same as the output neuron whose $PSP$ value would be maximal.

Specifically, output neuron $n_f$ firing as first is obtained as follows: initially, $\underline{PSP}_{n_i}$ of each output neurons in $\mathbf{NO}$ is reset to 0. Next, in the loop in which variable $j$ iterates over identifiers of input neurons starting from the one with the least order value (0) to the one with the greatest order value ($NI_{size}-1$), $\underline{PSP}_{n_i}$ of each output neuron $n_i$ in $\mathbf{NO}$ is calculated in an incremental way. As a result, after $k$ iterations, where $k \in \{1, 2, \dots NI_{size}\}$, $\underline{PSP}_{n_i}$ is equal to $w_{n_{j_0}n_i}\cdot mod^{order(j_0)} + w_{n_{j_1}n_i}\cdot mod^{order(j_1)} + \dots w_{n_{j_{k-1}}}n_i \cdot mod^{order(j_{k-1})}$, where $n_{j_l}$ is the input vector whose \textit{order} is equal to $l$, $l=0 \dots k-1$; that is,  $\underline{PSP}_{n_i} = w_{n_{j_0}n_i}\cdot mod^0 + w_{n_{j_2}n_i}\cdot mod^1 + \dots w_{n_{j_{k-1}}n_i}\cdot mod^{(k-1)}$, and $order(j_0) = 0$, $order(j_1) = 1$, \dots, $order(j_{k-1}) = k-1$.

After the first iteration, in which $\underline{PSP}$ (and by this, $PSP$) of at least one output neuron is greater than the $\gamma$ threshold, no other iterations are carried out. In such a case, each output neuron whose current $\underline{PSP}$ value is greater than $\gamma$ is added to the $\texttt{ToFire}$ list. $n_f$ is found as this output neuron in $\texttt{ToFire}$ that has the greatest value of $\underline{PSP}$, and is returned as the result of the $\Call{FiresFirst}{}$ function. Please note that the method we propose to calculate more and more precise lower approximations of $PSP$ of output neurons guarantees that $n_f$ is found in a minimal number of iterations.
If within $NO_{size}$ iterations no output neuron with $\underline{PSP} > \gamma$ is found, the $\Call{FiresFirst}{}$ returns NULL to indicate that no output neuron in $\mathbf{NO}$ was fired.

If $\Call{FiresFirst}{}$ returns NULL, then, in steps~12~to~14 of Algorithm~\ref{Alg:eSNN1}, value $x_t$ is classified as being anomalous and the prediction of network $y_t$ as well as error value $e_t$ are set to NULL and $+\infty$, respectively. Otherwise, in steps~16~to~18 of Algorithm~\ref{Alg:eSNN1}, the prediction of network $y_t$ is assigned output value $v_{n_f}$, error $e_t$ is set to the absolute difference between $x_t$ and $y_t$, and our proposed $\Call{ClassifyAnomaly}{}$ procedure is invoked.

$\Call{ClassifyAnomaly}{}$ is given in Algorithm~\ref{Alg:AnomalyClassification}. Its description was provided in subsection~\ref{subsubsec:AC}. The procedure returns  Boolean value $u_{t}$ indicating presence or absence of an anomaly for input value $x_{t}$.

% Fires First
\begin{algorithm}[h!t]
	\caption{\protect \Call{FiresFirst}{$CNO_{size}$}}
	\small{
	\begin{algorithmic}[1]
	    \Require $CNO_{size}$ - current size of output repository $\mathbf{NO}$
	    \Ensure $n_f$ - an output neuron $\in \mathbf{NO}$ which fires first
		%\Function{FiresFirst}{$CNO_{size}$}
            \State \texttt{ToFire} $\gets \emptyset$
            \State \begin{varwidth}[t]{\linewidth}
            %$\mathbf{SNI}$ $\gets $ sort $j \gets 1, \dots, NI_{size}$ according
            %\par
            %\hskip\algorithmicindent to ascending $order(j)$

            $\mathbf{SNIID}$ $\gets $ the list of identifiers of input neurons in $\mathbf{NI}$ obtained by sorting input neurons increasingly according to their $order$ value
            \end{varwidth}

            \For{$i \gets 0$ to $CNO_{size}-1$}
                    \State $\underline{PSP}_{n_i} \gets 0$
            \EndFor
            \For{$j \gets$ first to last input neuron identifier on
            \par \hskip\algorithmicindent
            list  $\mathbf{SNIID}$ }
            \par
                \For{$i \gets  0 $ to $CNO_{size}-1$} \Comment{output neuron ids}
                    \State $\underline{PSP}_{n_i} \gets \underline{PSP}_{n_i} + w_{n_jn_i} \cdot mod^{order(j)}$
                    \If{$\underline{PSP}_{n_i} > \gamma$}
                        \State Insert $n_i$ to \texttt{ToFire}
                    \EndIf
                \EndFor
                \If{\texttt{ToFire} $\neq \emptyset$}
                    \State \begin{varwidth}[t]{\linewidth}
                    $n_f \gets $ an output neuron in $\texttt{ToFire}$ such that
                    \par \hskip\algorithmicindent $\underline{PSP}_{n_f} =  max\{\underline{PSP}_{n_i}|  n_i \in \texttt{ToFire}\}$
                    %$n_f \gets n_i: (\underline{PSP}_{n_f} - \gamma) = $\par
                    %\hskip\algorithmicindent $max(\underline{PSP}_{n_i} - \gamma)$ for all $ n_i \in \texttt{ToFire}$
                    \end{varwidth}
                    \State \Return $n_f$
                \EndIf
            \EndFor
            \State \Return NULL
		%\EndFunction
	\end{algorithmic}
	}
	\label{Alg:FiresFirst}
\end{algorithm}

%Classify Anomaly

% Classify Anomaly
\begin{algorithm}[h!t]
	\caption{\protect \Call{ClassifyAnomaly}{$\mathbf{E}, \mathbf{U}$}}
	\small{
	\begin{algorithmic}[1]
	    \Require $\mathbf{E} = [e_1, \dots, e_{t}]$ - vector of error values; $\mathbf{U} = [u_1, \dots, u_{t - 1}]$ - vector of input values classified as anomalies or not; $e_t$ - error between predicted $y_t$ and input $x_t$ values.
	    \Ensure $u_{t}$ - a Boolean value being classification of $x_{t}$ as either an anomaly or not.
		%\Function{ClassifyAnomaly}{$\mathbf{E}, \mathbf{U}$}
		\State $\mathbf{e} \gets \emptyset $
		\State \begin{varwidth}[t]{\linewidth}
		Append to $\mathbf{e} $ all $e_k$ such that: \par
		\hskip\algorithmicindent $ k = t-(\mathcal{W}_{size} - 1), \dots, t - 1$ and $u_k$ is $False$
		\end{varwidth}
		\If{$\mathbf{e} = \emptyset$}
		\State $u_t = False$
		\Else
		\State Calculate $\overline{x}^{ }_{\mathbf{e}}$ and $s_{\mathbf{e}}$ over $\mathbf{e}$
		\If{$ {e_{t} - \overline{x}^{ }_{\mathbf{e}}} \geq \varepsilon \cdot s_{\mathbf{e}} $}
		\State $u_t = True$
		\Else
		\State $u_t = False$
		\EndIf
		\EndIf
		\State \Return $u_t$
		%\EndFunction
	\end{algorithmic}
	}
	\label{Alg:AnomalyClassification}
\end{algorithm}

In step~\ref{step:NeuronInit} of Algorithm~\ref{Alg:eSNN1}, new candidate output neuron $n_c$ is created, and then initialized in our proposed  $\Call{InitializeNeuron}{}$ procedure, which is presented in Algorithm~\ref{Alg:InitializeNeuron}. $\Call{InitalizeNeuron}{}$ first creates synapses between candidate output neuron $n_c$ and each input neuron in $\mathbf{NI}$. Then, the weights of the created synapses are calculated according to the firing order values of input neurons in $\mathbf{NI}$ obtained for input value $x_{t}$. Next, output value $v_{n_c}$ of $n_c$ is generated from a normal distribution created based on input values currently falling into window $\mathcal{W}$ (as it was presented in subsection~\ref{subsub:GenerationOutVal}), and finally the update time $\tau_{n_c}$ is set to current input time $t$. Additionally, if the anomaly is not detected ($u_{t}$ is $False$), then the value correction operation is performed which adjusts output value $v_{n_c}$ of the candidate output neuron $n_c$. Specifically, the value $v_{n_c}$ is increased or decreased by the factor $\xi \in [0, 1]$ of the difference $x_{t} - v_{n_c}$ (please see subsection~\ref{subsub:ValueCorr} for details).

%The initialization time $\tau_{n_i}$ of each output neuron $n_i \in \mathbf{NO}$ is used to determine if $n_i$ should be replaced with a new candidate neuron.

\begin{algorithm}[h!t]
	\caption{\protect \Call{InitializeNeuron}{$\mathcal{W}$}}
	\small{
	\begin{algorithmic}[1]
	    \Require $\mathcal{W}$ - current window of input values, $t$ - time of current input value $x_t$
	    \Ensure $n_c$ - a newly created and initialized candidate output neuron
		%\Function{InitializeNeuron}{$\mathcal{W}_t$}
		\State Create new neuron $n_c$
	    \For{$ j \gets 0$ to $NI_{size}-1$}
		    \State Create synapse between $n_j \in \mathbf{NI}$ and $n_c$
		 \EndFor
		 \For{$ j \gets 0$ to $NI_{size}-1$} \Comment{Calculate $\mathbf{w}_{n_c}$}
		    \State $w_{n_jn_c} \gets mod^{order(n_j)}$
		 \EndFor
		 %\State $PSP_{n_c}^{max} \gets 0$
		 %\For{$ j \gets 1$ to $NI_{size}$}
		 %  \State $PSP_{n_c}^{max} \gets w_{n_jn_c} \cdot mod^{order(j)} + PSP_{n_c}^{max}$
		 %\EndFor
        %\State $\gamma_{n_c} \gets PSP_{n_c}^{max} \cdot C$
		 \State $v_{n_c} \gets$ Generate output value from $\mathcal{N}(\overline{x}_{\mathcal{W}}^{ }, s_{\mathcal{W}}^2)$
		 \State $\tau_{n_c} \gets t$
		    \State $M_{n_c} \gets 1$
		    \State \Return $n_c$
		%\EndFunction
	\end{algorithmic}
	}
	\label{Alg:InitializeNeuron}
\end{algorithm}

In step~\ref{step:MinDistFor} of Algorithm~\ref{Alg:eSNN1}, the  $\Call{FindMostSimilar}{}$ procedure, presented in Algorithm~\ref{Alg:MostSimilarNeuron}, is called. The procedure finds an output neuron $n_s \in \mathbf{NO}$, such that the Euclidean distance $D_{n_c, n_s}$ between vectors of synapses weights of $n_c$ and $n_s$ is the smallest. If $D_{n_c,n_s}$ is less than or equal to the similarity threshold value $sim$, then $n_s$ is merged with $n_c$ according to the $\Call{UpdateNeuron}{}$ procedure, presented in Algorithm~\ref{Alg:UpdateNeuron}. The updated values of synapses weights, output value, update time and update counter of output neuron $n_s$  are calculated according  Eq.~(\ref{Eq:UpdateNeuron}).

Otherwise, if the number of output neurons in repository $\mathbf{NO}$ is still below $NO_{size}$, then $n_c$ is added to $\mathbf{NO}$ and counter $CNO_{size}$ is incremented.

If both the similarity condition is not fulfilled and the $\mathbf{NO}$ repository is full, then candidate output neuron $n_c$ replaces the oldest neuron $n_{oldest}$ in $\mathbf{NO}$ (that is, neuron $n_{oldest}$ in $\mathbf{NO}$ whose update time $\tau_{n_{oldest}}$ is minimal).

\begin{algorithm}[h!t]
	\caption{\protect \Call{FindMostSimilar}{$ n_c $} }
	\small{
	\begin{algorithmic}[1]
	    \Require $n_c$ - a candidate output neuron.
	    \Ensure $n_s$ - the neuron in $\mathbf{NO}$ such that Euclidean distance between $\mathbf{w}_{n_s}$ and $\mathbf{w}_{n_c}$ is least.
		%\Function{FindMostSimilar}{$n_c$}
		    \For {$i \gets 0 \dots CNO_{size}-1$}
		    \State {$D_{n_c, n_i} \gets  dist(\mathbf{w}_{n_c}, \mathbf{w}_{n_i})$}
		    \EndFor
		    \State \begin{varwidth}[t]{\linewidth}
		    $n_s \gets $ an output neuron in $\mathbf{NO}$ such that $D_{n_c, n_s} = $  \par
		    \hskip\algorithmicindent$min\{(D_{n_c, n_i})~|~i = 0, \dots, CNO_{size}-1\}$
		    \end{varwidth}
		    \State \Return $n_s$
		%\EndFunction
	\end{algorithmic}
	}
	\label{Alg:MostSimilarNeuron}
\end{algorithm}

\begin{algorithm}[h!t]
	\caption{\protect \Call{UpdateNeuron} {$n_s, n_c$}}
	\small{
	\begin{algorithmic}[1]
	    \Require $n_s$ - a neuron from $\mathbf{NO}$ to be updated; $n_c$ - a newly created candidate output neuron
		%\Procedure{UpdateNeuron}{$n_s, n_c$}
		    \State $\mathbf{w}_{n_s} \gets (\mathbf{w}_{n_c} + M_{n_s} \cdot \mathbf{w}_{n_s})/(M_{n_s}+1)$
		   % \State $\gamma_{n_s} \gets (\gamma_{n_c} + M_{n_s} \cdot \gamma_{n_s})/(M_{n_s}+1)$
		    \State $v_{n_s} \gets (v_{n_c} + M_{n_s} \cdot v_{n_s})/(M_{n_s}+1)$
		    \State $\tau_{n_s} \gets (\tau_{n_c} + M_{n_s} \cdot \tau_{n_s})/(M_{n_s}+1)$
		    \State $M_{n_s} \gets M_{n_s} + 1$
		    \State \Return
		%\EndProcedure
	\end{algorithmic}
	}
	\label{Alg:UpdateNeuron}
\end{algorithm}

%\newpage
\section{Experiments}
\label{sec:Experimental Results}

In this section, we present the results of the comparative experimental evaluation of the proposed OeSNN-UAD method and state-of-the-art methods and algorithms for unsupervised anomaly detection. For comparison, we use the following  methods and algorithms: Numenta \citep{Ahmad2017}, NumentaTM \citep{Ahmad2017}, HTM JAVA \citep{Hawkins2016}, Skyline \citep{Skyline}, TwitterADVec \citep{TwitterAD2015}, Yahoo EGADS (Extensible Generic Anomaly Detection System) \citep{Laptev2015}, DeepAnT \citep{Munir2019_IEEEAccess_DEEPAnT}, Bayesian Changepoint \citep{Adams2007_CoRR_BayesChangePT}, EXPected Similarity Estimation (EXPoSE) \citep{Schneider2016}, KNN CAD \citep{Burnaev2016}, Relative Entropy \citep{Wang2011} and ContextOSE \citep{Contextual} and the unsupervised anomaly detection method offered in \citep{Zhang2019-AnomalyDetectionSliding}. The experiments were carried out on two anomaly benchmark repositories: Numenta Anomaly Benchmark \citep{NAB2019} and Yahoo Anomaly Dataset \citep{YahooAnomalyDataset}. The experimental results concerning both our proposed OeSNN-UAD and all other anomaly detectors used for comparative assessment were obtained after tuning their parameters.

The section starts with an overview of both of the aforementioned benchmark repositories. Then, it is followed by the description of the experimental setup. The OeSNN-UAD parameter tuning method is presented as well. Next, we present the extensive experimental evaluation of the compared methods and algorithms is provided. Finally, we provide and discuss the results of the experiments examining influence of different values of $\mathcal{W}_{size}$ and $\varepsilon$ parameters on the quality of the anomaly detection.

\subsection{Anomaly Benchmark Repositories Used in the Experiments}

Both repositories (Numenta Anomaly Benchmark and Yahoo Anomaly Dataset), which we use for experiments, contain time series with labeled anomalies. Both of them are commonly used to asses the quality of unsupervised anomaly detection for various methods and algorithms (please see, for example \citep{Ahmad2017,Munir2019_IEEEAccess_DEEPAnT,Zhang2019-AnomalyDetectionSliding}).

\subsubsection{The Numenta Anomaly Benchmark Repository}
\label{subsec:RepositoriesNAB}

The Numenta Anomaly Benchmark (NAB) repository contains 7 categories of datasets, both artificial and real, each of which has multiple CSV data files. Each CSV data file consists of two time series, one of them being a series of timestamp values and the second one being a series of input values. The number of input values in data files varies between \num{1000} and \num{22000}. Overall, there are 58 data files in NAB. All time series in the NAB repository are imbalanced with the average percentage of input values being anomalies in a time series less than 10\% on average. In the current version (1.0) of NAB, the following categories of data files are distinguished:
\begin{itemize}
\item \texttt{artificialNoAnomaly} - contains data files artificially generated, which do not have anomalies;
\item \texttt{artificialWithAnomaly} - contains data files which consist of artificial data with anomalies;
\item \texttt{realAdExchange} - contains data files with online advertisements clicks recordings;
\item \texttt{realAWSCloudwatch} - contains data files with metrics from AWS servers;
\item \texttt{realKnownCauses} - contains real data files, such as hourly registered taxi schedules in New York City or CPU utilization;
\item \texttt{realTraffic} - contains data files with freeway traffic recordings, such as speed or travel time;
\item \texttt{realTweets} - contains data files with Tweeter volume statistics.
\end{itemize}

\begin{figure}[h!t]
	\centering
	\includegraphics[width=0.7\linewidth]{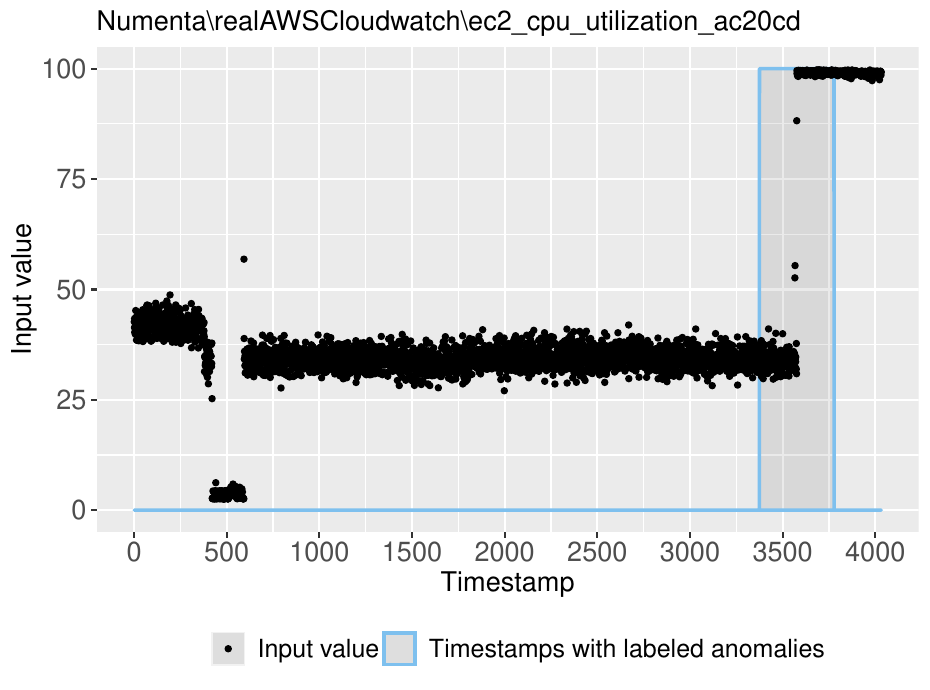}
	\caption{Input values and labeled anomalies in data file \texttt{ec2\_cpu\_utilization\_ac20cd} of \texttt{RealAWSCloudwatch} category in Numenta Anomaly Benchmark. Anomalous input values, which occur around timestamp 500 are not labeled as such, whereas anomaly window around timestamp 3500 incorrectly identifies input values as anomalies.}
	\label{Fig:NumentaAnomaliesExample}
\end{figure}

Only data files in \texttt{artificialNoAnomaly} category do not contain anomalies. The data files in the remaining categories contain at least one anomaly window. Each anomaly window consists of multiple input values and each data file can have several anomaly windows. The labeling of anomaly windows in the data files was conducted manually or by means of algorithms \citep{NAB2019-labeling}. However, as follows from the documentation of NAB, it is not guaranteed that all anomalies in a data file are labeled \citep{NAB2019-labeling}. In fact, potential users of NAB are encouraged to perform additional anomaly labeling of data files \citep{NAB2019-labeling}, which can be released in future versions of NAB. It was also reported in \citep{Singh2017} that some data files in NAB contain missing values or differences in input values distributions. For these reasons, NAB is particularly challenging for anomaly detection algorithms. In Figure~\ref{Fig:NumentaAnomaliesExample}, we illustrate data file \texttt{ec2\_cpu\_utilization\_ac20cd} from \texttt{realAWSCloudwatch} category. It can be noted that input values around timestamp 500, which can intuitively be perceived as anomalous, are not labeled as such, whereas all input values in window starting at timestamp 3375 and ending at timestamp 3777, which are labeled in the dataset as anomalies, in the majority of cases seem not to be anomalous. This kind of incorrect anomaly labeling or its lack can negatively influence the measures describing anomaly detection quality, as well as make learning of a detector less effective.

\subsubsection{The Yahoo Anomaly Dataset Repository}

The Yahoo Anomaly Dataset \citep{YahooAnomalyDataset} repository consists of four categories of data files:

\begin{itemize}
    \item \texttt{A1Benchmark} - contains 67 data files with real input time series values. Both single anomalous values and windows of anomalies occur in these data files. Each data file consists of three time series: timestamps, input values and a label for each input value as being either anomalous or not. %As the authors of the repository noted \citep{YahooAnomalyDataset}, anomaly labeling in the case of some data files of this category can be inaccurate.
    \item \texttt{A2Benchmark} - consists of 100 synthetic data files, which contain anomalies in the form of single anomalous values. Most of input time series values in this category have their own periodicity. Similarly to \texttt{A1Benchmark}, each data file contains only three time series: timestamp, input values and labels indicating the presence or absence of anomalies.
    \item \texttt{A3Benchmark} - has 100 synthetic data files with anomalies in the form of single anomalous values. In comparison to \texttt{A2Benchmark}, input values time series in this category are more noisy. In addition to three standard time series (timestamps, input values and anomalies labels), data files in this category contain also other time series (trend, noise, seasonality and changepoint), which are not used by our OeSNN-UAD model neither for anomaly detection nor for the calculation of the detection quality.
    \item \texttt{A4Benchmark} - contains 100 synthetic data files with anomalies. The majority of the anomalies correspond to sudden transitions from an input data trend to another significantly different input data trend.  \texttt{A4Benchmark} consists of the same time series types as \texttt{A3Bechnmark}. Again, in our detector we used only first three time series in each data file: timestamp, input values and anomalies labels. The former two are used for anomaly detection by OeSNN-UAD, while the third time series is used for calculation of its detection quality.
\end{itemize}

All time series in the Yahoo repository are imbalanced or strongly imbalanced with the average percentage of input values being anomalies in a time series less than 1\% on average. Similarly to the NAB repository, data files in the Yahoo Anomaly Dataset repository are provided as CSV files. %The anomalies detection quality using our approach on the Yahoo Anomaly Dataset repository is compared with TwitterADVec and DeepAnT algorithms as well as with Yahoo EGADS algorithm given in \citep{Laptev2015}.

\subsection{Experimental Setup and Optimization}
\label{subsec:experimentalsetup}

To run our OeSNN-UAD detector one has to provide values of the following parameters: $NI_{size}$ - number of input neurons,  $NO_{size}$ - the maximal number of output neurons, $mod$ - modulation factor, $C$ - fraction of $PSP^{max}$ required to fire an output neuron, $sim$ - threshold value for similarity between a candidate output neuron and an output neuron in terms of their weights vectors, which is required for merging these neurons, $\xi$ - output value correction factor, $W_{size}$ - window size,  $\varepsilon$ - anomaly classification factor.
The values of the $\beta$ and $TS$ parameters do not need to be determined, because the classification result does not depend on them (please see, Proposition~\ref{Proposition:Encoding}.(v)).
In the reported experiments, values of the first 6 parameters were set as follows: $NI_{size} = 10$, $NO_{size} = 50$, $sim = 0.17$, $mod = 0.6$, $C = 0.6$, $\xi = 0.9$. In our preliminary experiments, we observed that values of parameters $\mathcal{W}_{size}$ and $\varepsilon$ have the greatest impact on anomaly detection (similar observation follows from the experiments reported  in \citep{Ahmad2017,Munir2019_IEEEAccess_DEEPAnT}, in which approaches to anomaly detection that also use window size and anomaly classification threshold are proposed). Hence, the selection of $\mathcal{W}_{size}$ and $\varepsilon$ parameters values of OeSNN-UAD was optimized. In the case of data files from the Numenta Anomaly Benchmark repository, OeSNN-UAD was run multiple times with $W_{size} \in \{100, 200, \dots, 600\}$ and $\varepsilon \in \{2, 3, \dots, 7\}$, while in the case of data files from the Yahoo Anomaly Datasets repository, it was executed with  $W_{size} \in \{20, 40, \dots, 500\}$ and $\varepsilon \in \{2, 3, \dots, 17\}$.

Following \citep{Kasabov2014,Tu2017,Maciag2019_AirPollutionESNN}, we implemented the grid search procedure to find the best values of parameters $W_{size}$ and $\varepsilon$ for each data file separately. The grid search procedure iterates over all given combinations of input learning parameters to find a set of parameters values (in particular, $W_{size}$ and $\varepsilon$), which provides the best anomaly detection results for an input data file.

%Such an approach to grid search over learning parameters is motivated by the following objectives:
%\begin{itemize}
%    \item A similar grid search procedure is used for data files from the Numenta Anomaly Benchmark and Yahoo Anomaly Dataset repositories in previous works, such as \citep{Ahmad2017}.
%    \item In our approach, we implemented the grid search procedure to find the best values of parameters $W_{size}$ and $\varepsilon$. The grid search procedure iterates over all given combinations of input learning parameters to find a set of parameters (in particular, $W_{size}$ and $\varepsilon$), which provides the best anomaly detection results for an input data file.
%Such an approach to grid search or instance, for \texttt{realKnownCause}, which contain data files with input values from different domains such as NYC taxi traffic or CPU utilization;
    %\item for most of the data files in both repositories it is not possible to divide time series of each data file into a validation part (which can be possibly used for parameters tuning) and a test part (which is used to obtain quality of anomaly detection with the parameters that give the best anomaly detection results on the validation part). This is because the input time series usually contain only single anomalous values, and thus splitting the data file into validation and test parts can result in incorrectly labeled anomalies or the lack of them in either part.
% \end{itemize}
%Our experiments were performed on a computer equipped with Intel Core i7-8750H CPU and 16.0 GB of RAM memory.

The implementation of OeSNN-UAD is prepared in C++ and its source code  is publicly available (https://github.com/piotrMaciag32/eSNN-AD). The compiled executable file is lightweight (it consumes around 2 MB of RAM memory), which makes it additionally suitable for environments with very strict memory constraints, such as sensor microcontrollers or IoT devices.

% \begin{figure*}[h!t]
% 	\centering
% 	\includegraphics[width=1.0\linewidth]{n_awa_art_daily_jumpsup.pdf}
% 	\includegraphics[width=1.0\linewidth]{n_rae_exchange_4_cpm.pdf}
%     \includegraphics[width=1.0\linewidth]{n_rt_speed_7578.pdf}
%     \includegraphics[width=1.0\linewidth]{y_a1_real_19.pdf}
% 	\caption{The results of anomaly detection with OeSNN-UAD for four example data files in Numenta Anomaly Benchmark and Yahoo Anomaly Dataset.}
% 	\label{Fig:Plots}
% \end{figure*}

\subsection{Obtained Anomaly Detection Results}
\label{subsec:Results}

In the experimental phase, we compare anomaly detection quality of our approach to the other state-of-the-art methods and algorithms provided in the literature. To this end, we use five measures of detection quality: \textit{precision}, \textit{recall}, \textit{F-measure}, \textit{balanced accuracy} (BA) and \textit{Matthews correlation coefficient} (MCC).

\textit{Precision} provides information on how many of the input values detected as anomalies by the detector are actually labeled as anomalies in data files, while \textit{recall} indicates how many of the labeled anomalies in the data file are properly detected by the detector. \textit{F-measure} (F1) is a harmonic mean of precision and recall. Moreover, since most of the datasets in the NAB and Yahoo repositories are strongly imbalanced, we additionally computed \textit{balanced accuracy}. \textit{Balanced accuracy} (BA) is defined as the average of recall and the equivalent of recall calculated with respect to the category of non-anomalous input values. This measure is typically used when dealing with imbalanced datasets, such as time series in the NAB and Yahoo repositories. \textit{Matthews correlation coefficient} (MCC) is defined as a correlation coefficient between real and predicted labels of both anomalous and non-anomalous input values.

In Eqs.~(\ref{Eq:Precision}),~(\ref{Eq:Recall}),~(\ref{Eq:F-measure}),~(\ref{Eq:BA}),~(\ref{Eq:MCC}) we give formulae for precision, recall, F-measure, balanced accuracy and Matthews correlation coefficient. In these equations, |TP| (True Positives) denotes the number of input values that were both classified as anomalies by the detector and labeled as being such in the data file, |FP| (False Positives) denotes the number of input values that were classified as  anomalous by the detector, but were not labeled as anomalies in the data file, while |FN| (False Negatives) denotes the number of input values labeled as anomalous in the data file, but not classified as anomalies by the detector.
\begin{equation}
    \text{Precision} = \frac{\text{|TP|}}{\text{|TP|} + \text{|FP|}}.
    \label{Eq:Precision}
\end{equation}
\begin{equation}
    \text{Recall} = \frac{\text{|TP|}}{\text{|TP|} + \text{|FN|}}.
    \label{Eq:Recall}
\end{equation}
\begin{equation}
    \text{F1} = 2\cdot \frac{\text{Precision}\cdot \text{Recall}}{\text{Precision} + \text{Recall}}.
    \label{Eq:F-measure}
\end{equation}
\begin{equation}
    \text{BA} = \frac{1}{2} \cdot \Big(\frac{|TP|}{|TP|+|FN|} + \frac{|TN|}{|TN|+|FP|}\Big).
    \label{Eq:BA}
\end{equation}
\begin{equation}
    \text{MCC} = \frac{|TP|\cdot |TN| - |FP|\cdot |FN|}{\sqrt{(|TP|+|FP|)(|TP|+|FN|)(|TN|+|FP|)(|TN|+|FN|)}}.
    \label{Eq:MCC}
\end{equation}

\begin{figure*}[h!t]
	\centering
	\includegraphics[width=1.0\linewidth]{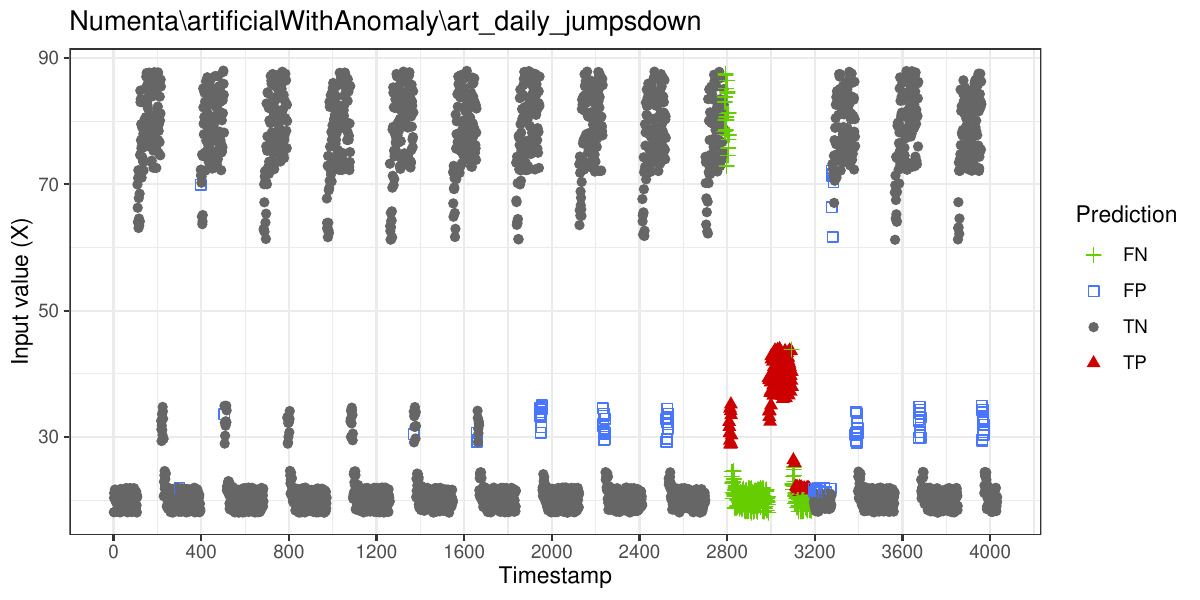}
	\includegraphics[width=1.0\linewidth]{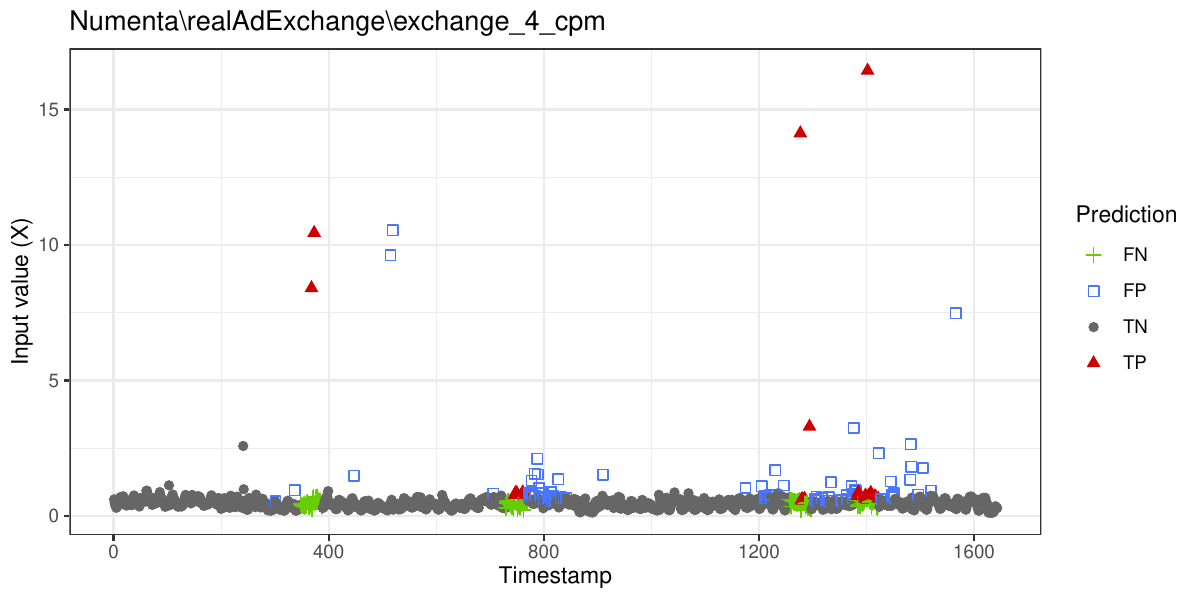}
    \includegraphics[width=1.0\linewidth]{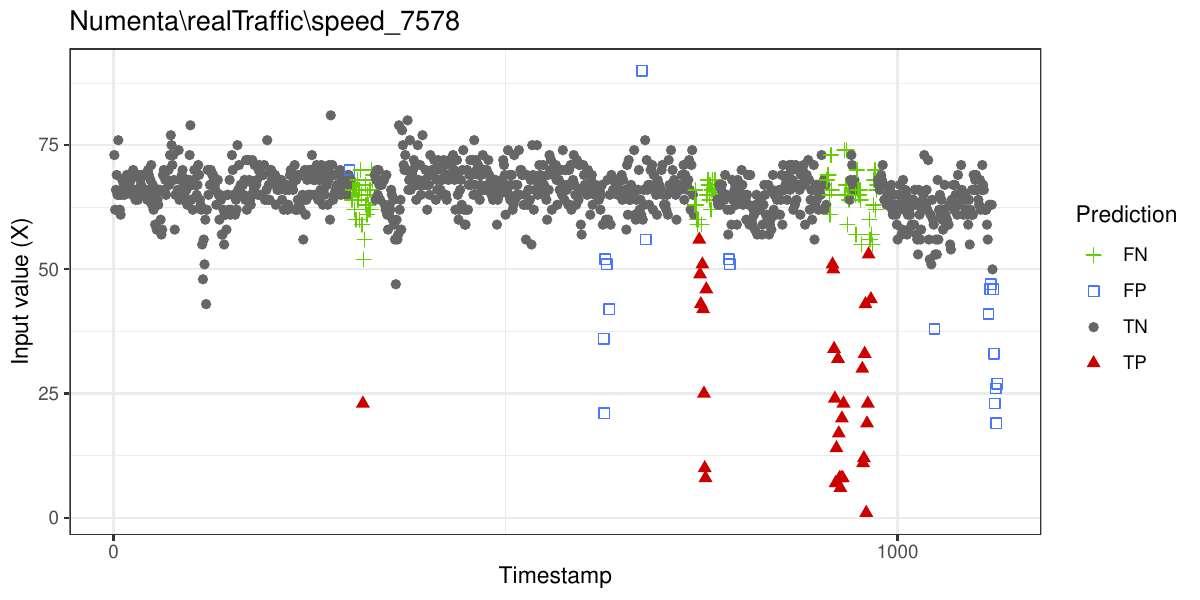}
	\caption{Anomaly detection results for selected data files from the Numenta Anomaly Benchmark repository with OeSNN-UAD.}
	\label{Fig:PlotsPredN}
\end{figure*}

\begin{figure*}[h!t]
	\centering
	\includegraphics[width=1.0\linewidth]{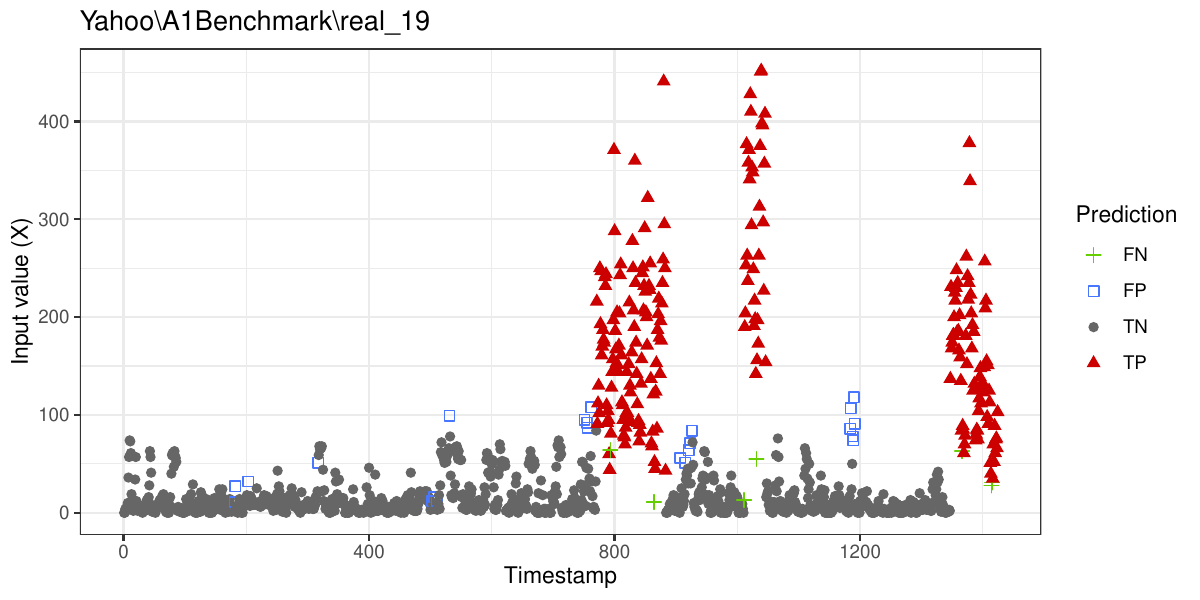}
	\includegraphics[width=1.0\linewidth]{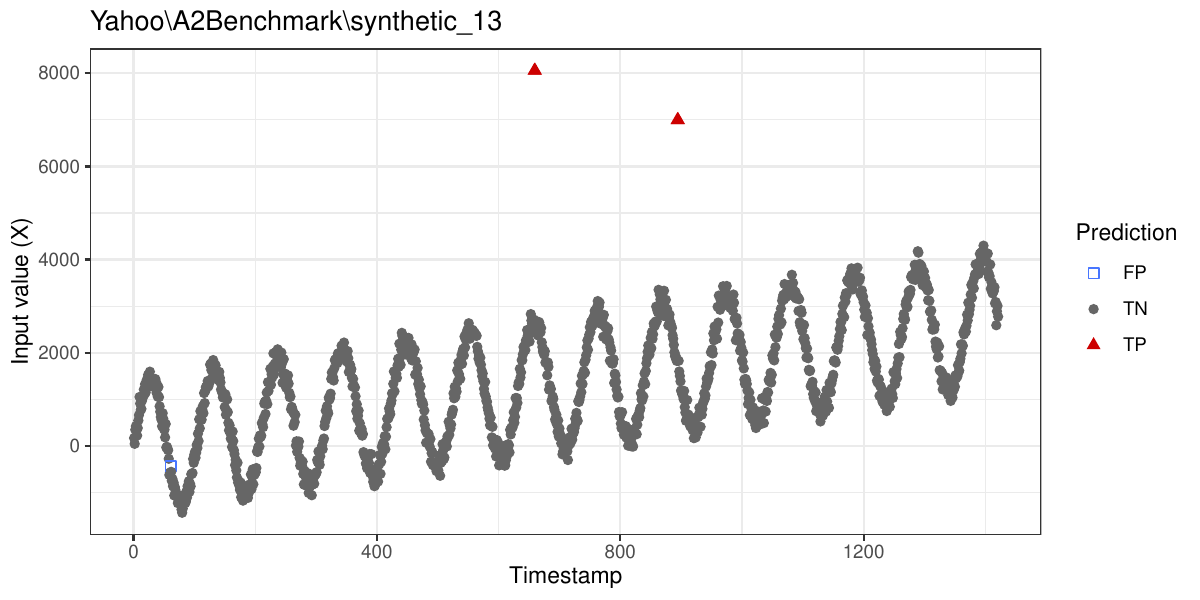}
    \includegraphics[width=1.0\linewidth]{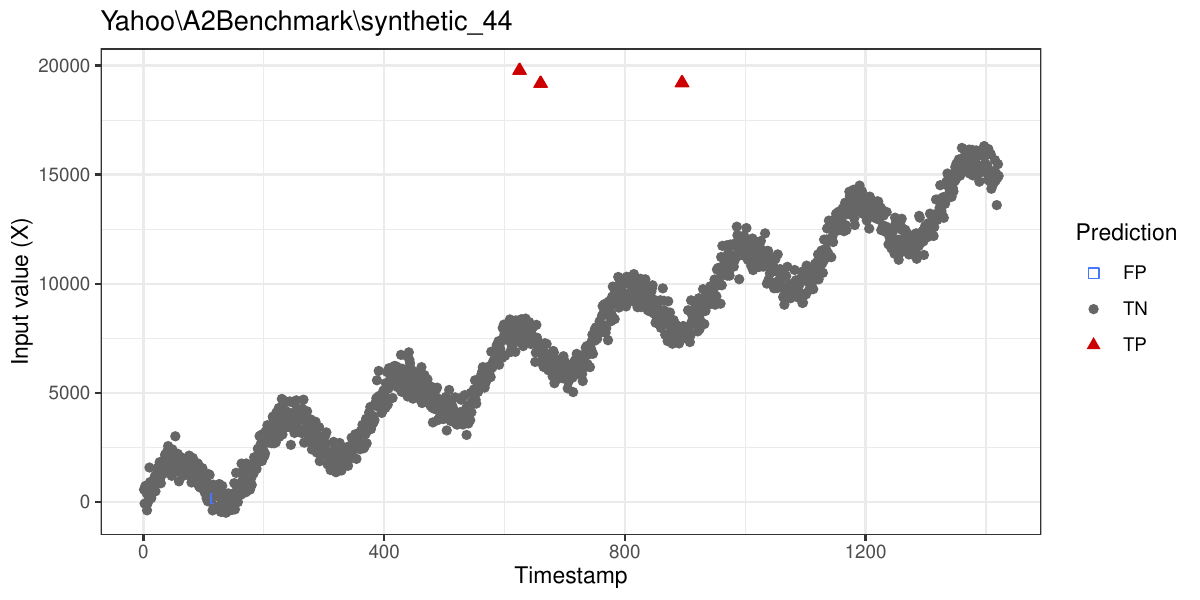}
	\caption{OeSNN-UAD anomaly detection results for selected Yahoo data files. %from the Yahoo Anomaly Benchmark repository%
	The red points on the chart \texttt{real\_19} present collective anomalies (timestamp $\sim$1000) as well as both collective and contextual anomalies (timestamps
	$\sim$800 and $\sim$1400); the red points on the other two charts present point anomalies.}
	\label{Fig:PlotsPredY}
\end{figure*}

In Figure~\ref{Fig:PlotsPredN}, we present charts showing anomaly detection results obtained for example data files in categories \texttt{artificialWithAnomaly, realAdExchange} and \texttt{realTraffic} of the NAB repository. Figure~\ref{Fig:PlotsPredY} shows similar charts for example datasets in categories \texttt{A1Benchmark}, \texttt{A2Benchmark} of the Yahoo Anomaly Dataset repository. The charts present the occurrences of true positives (TP), false positives (FP), false negatives (FN) and true negatives (TN) found by the proposed OeSNN-UAD algorithm. It can be observed on the charts, especially in the case of the data files from the NAB repository, that there are some input values in these data files that are not correctly labeled as anomalies. For example, it can be seen on the first chart of Figure~\ref{Fig:PlotsPredN}, which presents anomaly detection in \texttt{art\_daily\_jumpsdown} data file from \texttt{artificialWithAnomaly}, that many input values that occur around time period 2800 - 3200 are labeled in the data file as anomalies, while they should be perceived as non-anomalous (please see the green groups of input values). In this case, OeSNN-UAD classifies them, as we believe, correctly as non-anomalous, but they are treated as false negatives with respect to available labeling of the data file. On the other hand, it may happen that true anomalous input values are not labeled as anomalies in these data files. Such a phenomenon may result in the decrease in the value of the detection quality measures, such as F-measure, even though a detector correctly classifies input data as anomalous or non-anomalous, respectively.

Our proposed OeSNN-UAD is able to detect point anomalies (see e.g. Figure~\ref{Fig:PlotsPredY}: \texttt{synthetic\_13} and \texttt{synthetic\_44}), contextual anomalies and collective anomalies (see e.g. Figure~\ref{Fig:PlotsPredY}:\texttt{real\_19}).\footnote{The characteristics of point, collective and contextual types of anomalies can be found in \citep{Chandola2009}.} Proper detection of these types of anomalies, however, is often dependent on the used values of parameters (especially on window size $\mathcal{W}_{size}$ and anomaly factor $\varepsilon$). When it is possible, one should adjust the values of these parameters based on analysis of a subset of available data stream.

% \begin{landscape}
% \begin{figure*}[h!t]
% 	\centering
%     \includegraphics[width=1.0\linewidth]{rev_exp1_v3.pdf}
% 	\caption{The modified.}
% 	\label{Fig:PlotsModifiedDataset}
% \end{figure*}
% \end{landscape}

In Table~\ref{Table:ResultsNAB}, we show the results of OeSNN-UAD anomaly detection for the Numenta Anomaly Benchmark repository as well as for the other unsupervised anomaly detection methods and algorithms. As in  \citep{Munir2019_IEEEAccess_DEEPAnT}, we report the mean F-measure obtained for each category of data files for each compared detector. As follows from Table~\ref{Table:ResultsNAB}, OeSNN-UAD outperforms the results obtained by the other detectors in terms of F-measure for each category of data files.

\begin{landscape}
\begin{figure*}[h!t]
	\centering
	\includegraphics[width=1.0\linewidth]{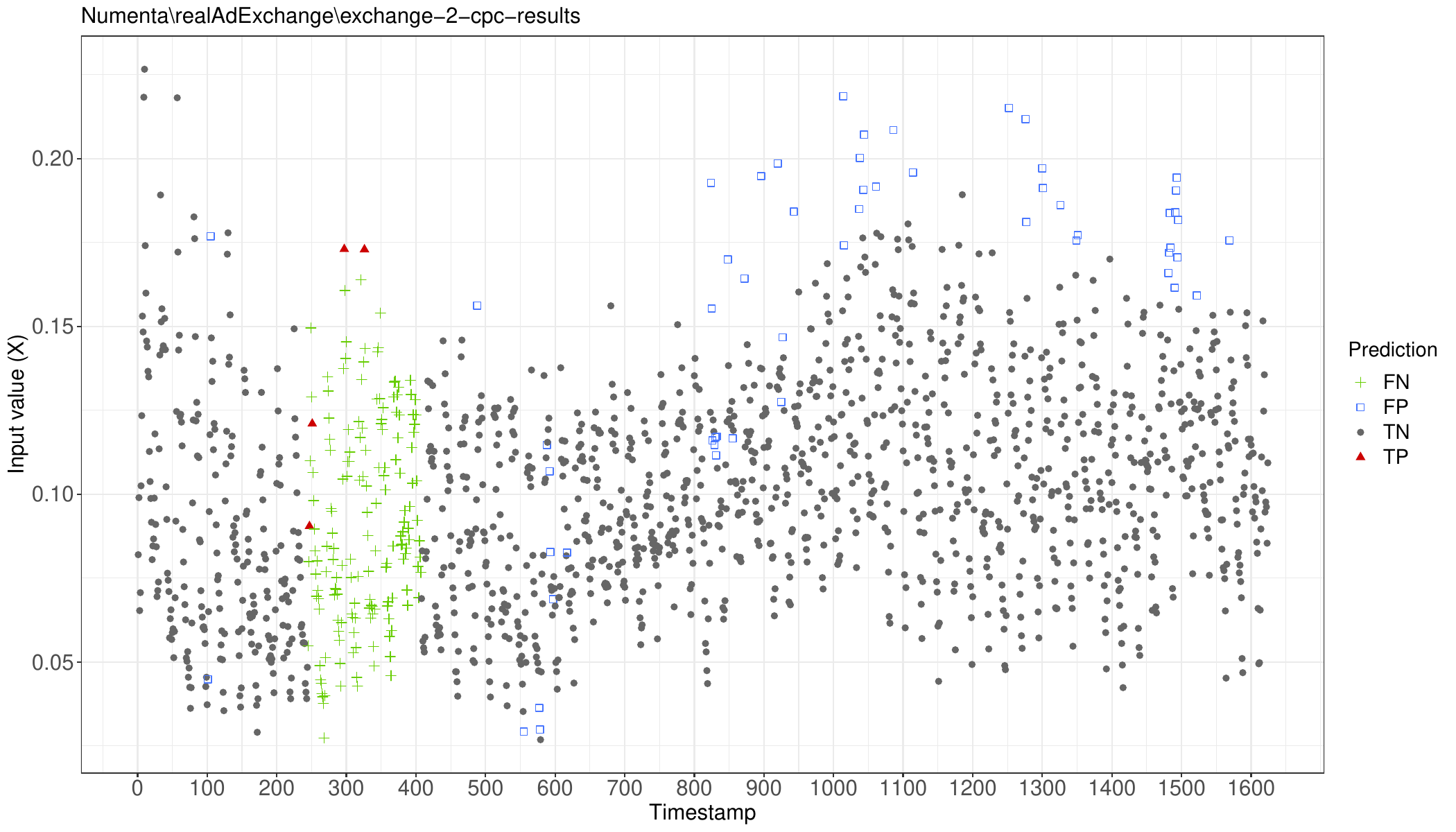}
	\caption{Anomaly detection for data file \texttt{exchange-2-cpc-results} in \texttt{realAdExchange} category of data files of NAB with OeSNN-UAD.}
	\label{Fig:NumExchange2CPCRes}
\end{figure*}
\end{landscape}

Table~\ref{Table:ResultsNABDatasets} presents the obtained precision and recall values for the selected data files from the Numenta Anomaly Benchmark repository. For some data files, OeSNN-UAD is able to provide much higher values of both precision and recall than the other detectors. Most of the detectors compared with OeSNN-UAD, for which the results are presented in Table~\ref{Table:ResultsNABDatasets}, have very high values of precision, but very low values of recall. High values of precision imply that very few cases which were not labeled as anomalies are detected as anomalous. Low recall, on the other hand, implies that these detectors do not discover a large number of cases labeled as anomalies. In fact, the recall values below $0.01$ obtained for many data files presented in Table~\ref{Table:ResultsNABDatasets} by these detectors indicate their very limited ability to detect anomalies. To the contrary, our OeSNN-UAD detector, in the majority of cases, is characterized by much larger values of the recall, and thus it is much more efficient in detecting anomalies than the compared detectors. The only case, when OeSNN-UAD has low recall value can be observed for \texttt{exchange-2-cpc-results} data file. Nevertheless, as follows from Figure~\ref{Fig:NumExchange2CPCRes}, which presents this data file and the results of anomaly detection obtained with OeSNN-UAD, a number of input values in this data file were labeled in a counter-intuitive way. For example, the majority of input values in the time interval 245-407 were found by OeSNN-UAD as non-anomalous (namely, input values marked as green, which seem not to be real anomalies). Surprisingly, these values are labeled as anomalies in the data file. This incoherence between given labeling of input values and their expected labeling results in increase in the number of false negatives and, in consequence, in decrease in the recall value for OeSNN-UAD. In addition, high input values (marked as blue ones) in the time interval 1000-1600 were found by OeSNN-UAD as anomalies (and seem to be real anomalies), but are not labeled as such in the data file. This results in an increase in the number of false positives and, in consequence, in a decrease in the precision value for OeSNN-UAD.

In Table~\ref{Table:NABDatasetsParameters}, we present the obtained optimal values of $\mathcal{W}_{size}$ and anomaly classification factor $\varepsilon$ of OeSNN-UAD for data files, which were used in the experiments the results of which are presented in Table~\ref{Table:ResultsNABDatasets}.

In Table~\ref{Table:ResultsYahoo}, we present the comparison of the obtained F-measure values for each category of data files in the Yahoo Anomaly Dataset repository. For the real data files category (\texttt{A1Benchmark}), the proposed OeSNN-UAD approach provides higher values of the F-measure than the recent results reported in the literature \citep{Munir2019_IEEEAccess_DEEPAnT}, while for the other three categories of data files, OeSNN-UAD's results are competitive to the results reported there.

\begin{landscape}
\begin{table*}[h!t]
\centering
\caption{Average F-measure values obtained for Numenta Anomaly Benchmark stream data using the unsupervised anomaly detectors (marked with *) presented in \citep{Munir2019_IEEEAccess_DEEPAnT} and using our proposed OeSNN-UAD detector. The bolded results are the best for each data files category. The results for the detectors marked with * were reported in  \citep{Munir2019_IEEEAccess_DEEPAnT}.}
\begin{tabular}{llllllllllllll}
	\toprule
     Dataset category & \parbox[t]{2mm}{{\rotatebox[origin=l]{80}{Bayesyan Changepoint*}}} & \parbox[t]{2mm}{{\rotatebox[origin=l]{80}{Context OSE*}}} & \parbox[t]{2mm}{{\rotatebox[origin=l]{80}{EXPoSE*}}} & \parbox[t]{2mm}{{\rotatebox[origin=l]{80}{HTM JAVA*}}} & \parbox[t]{2mm}{{\rotatebox[origin=l]{80}{KNN CAD*}}} & \parbox[t]{2mm}{{\rotatebox[origin=l]{80}{Numenta*}}} & \parbox[t]{2mm}{{\rotatebox[origin=l]{80}{NumentaTM*}}} & \parbox[t]{2mm}{{\rotatebox[origin=l]{80}{Relative Entropy*}}} & \parbox[t]{2mm}{{\rotatebox[origin=l]{80}{Skyline*}}} & \parbox[t]{2mm}{{\rotatebox[origin=l]{80}{Twitter ADVec*}}} & \parbox[t]{2mm}{{\rotatebox[origin=l]{80}{Windowed Gaussian*}}} & \parbox[t]{2mm}{{\rotatebox[origin=l]{80}{DeepAnT*}}} & \parbox[t]{2mm}{{\rotatebox[origin=l]{80}{OeSNN-UAD}}} \\
    \midrule
     Artificial no Anomaly& 0 & 0 & 0 & 0 & 0& 0 & 0 & 0 & 0 & 0 & 0 & 0 & 0 \\
     \midrule
     Artificial with Anomaly& 0.009  & 0.004 & 0.004 & 0.017 & 0.003 & 0.012  & 0.017 & 0.021 & 0.043 & 0.017 & 0.013 & 0.156 & \textbf{0.427} \\
     \midrule
     Real Ad Exchange& 0.018 & 0.022 & 0.005 & 0.034  & 0.024 & 0.040 & 0.035 & 0.024 & 0.005 & 0.018 & 0.026 & 0.132 & \textbf{0.234}\\
     \midrule
     Real AWS Cloud & 0.006 & 0.007 & 0.015 & 0.018  & 0.006 & 0.017 & 0.018 & 0.018 & 0.053 & 0.013 & 0.06 & 0.146 & \textbf{0.369} \\
     \midrule
     Real Known Cause& 0.007 & 0.005 & 0.005 & 0.013 & 0.008 & 0.015 & 0.012 & 0.013 & 0.008 & 0.017 & 0.006 & 0.2 & \textbf{0.324} \\
     \midrule
     Real Traffic& 0.012  & 0.02 & 0.011 & 0.032 & 0.013 & 0.033 & 0.036 & 0.033 & 0.091 & 0.020 & 0.045 & 0.223 & \textbf{0.340}\\
     \midrule
     Real Tweets & 0.003 & 0.003 & 0.003 & 0.010 & 0.004 & 0.009 & 0.010 & 0.006 & 0.035 & 0.018 & 0.026 & 0.075 & \textbf{0.310}\\
	\bottomrule
\end{tabular}
\label{Table:ResultsNAB}
\end{table*}
\end{landscape}

\begin{landscape}
\begin{table*}[h!t]
\centering
\caption{Precision and recall values obtained for Numenta Anomaly Benchmark stream data using the selected unsupervised anomaly detectors (marked with *) presented in \citep{Munir2019_IEEEAccess_DEEPAnT} and using our proposed OeSNN-UAD detector. The results for the detectors marked with * were reported in \citep{Munir2019_IEEEAccess_DEEPAnT}.}
\small{
\begin{tabular}{ll|ll|ll|ll|ll|ll|ll}
	\hline
    \multicolumn{2}{c|}{\multirow{2}{*}{Time series}} & \multicolumn{2}{l|}{\makecell[l]{ContextOSE*}} &  \multicolumn{2}{l|}{\makecell[l]{NumentaTM*}} & \multicolumn{2}{l|}{\makecell[l]{Skyline*}} & \multicolumn{2}{l|}{\makecell[l]{ADVec*}} & \multicolumn{2}{l|}{\makecell[l]{DeepAnT*}} & \multicolumn{2}{l}{\makecell[l]{OeSNN-UAD}} \\
    & & Prec. & Rec. & Prec. & Rec. & Prec. & Rec. & Prec. & Rec. & Prec. & Rec. & Prec. & Rec.\\
    \hline\hline
    \multirow{2}{*}{\makecell[l]{Real AWS \\Cloud Watch}} & \makecell[l]{ec2-cpu-utilization-5f5533} & 1 & 0.005  & 1 & 0.01 & 1 & 0.002 & 1 & 0.002 & 1 & 0.01 & 0.18 & 0.51 \\\cline{2-14}
    & \makecell[l]{rds-cpu-utilization-cc0c53} & 1 & 0.005 & 1 & 0.002 & 1 & 0.1 & 0.62 & 0.012 & 1 & 0.03 & 0.50 & 0.75 \\
    \hline
    \hline
    \multirow{11}{*}{\makecell[l]{Real Known\\ Cause}} & \makecell[l]{ambient-temperature-\\system-failure} &  0.33 & 0.001  & 0.5 & 0.006 & 0  & 0 & 0 & 0 & 0.26 & 0.06 & 0.21 & 0.75 \\\cline{2-14}
    & \makecell[l]{cpu-utilization-asg-\\misconfiguration}&0.12 &0.001  &0.52  &0.01  &0  &0  &0.74 &0.01  &0.63  &0.36  & 0.32 & 0.49 \\\cline{2-14}
    & \makecell[l]{ec2-request-latency-\\system-failure} &1  &0.009  &1  &0.009  &1  &0.014 &1  &0.02  &1  &0.04  & 0.38 & 0.40  \\\cline{2-14}
    & \makecell[l]{machine-temperature-\\system-failure} &1  &0.001  &0.27  &0.004  &0.97  &0.01 &1  &0.02  &0.8  &0.001  & 0.39 & 0.50 \\\cline{2-14}
    & \makecell[l]{nyc-taxi} &1  &0.002  &0.85  &0.006  &0  &0 &0  &0  &1  &0.002  & 0.17 & 0.47 \\\cline{2-14}
    & \makecell[l]{rouge-agent-key-hold} &0.33  &0.005  &0.5  &0.005  &0  &0 &0  &0  &0.34  &0.05  & 0.13 & 0.23 \\\cline{2-14}
    & \makecell[l]{rouge-agent-key-updown} &0  &0  &0  &0  &0  &0 &0.11  &0.002  &0.11  &0.01  & 0.25 & 0.43 \\
      \hline
      \hline
    \multirow{7}{*}{\makecell[l]{Real Traffic}}
    & \makecell[l]{occupancy-6005} &0.5  &0.004   &0.2  &0.004 &0.5  &0.004 &0.5  &0.004  &0.5  &0.004  &0.18  &0.41  \\ \cline{2-14}
    & \makecell[l]{occupancy-t4013} &1&0.008  &0.66  &0.008  &1  &0.04  &1 &0.02  &1  &0.036  &0.50  &0.44    \\\cline{2-14}
    & \makecell[l]{speed-6005} &0.5  &0.004  &0.25  &0.008  &1  &0.01 &1  &0.01  &1  &0.008  &0.36  &0.34  \\\cline{2-14}
    & \makecell[l]{speed-7578} &0.57  &0.03  &0.6  &0.02  &0.86  &0.16 &1  &0.01  &1  &0.07  &0.64  &0.30  \\\cline{2-14}
    & \makecell[l]{speed-t4013} &1  &0.008  &0.8  &0.01  &1  &0.06 &1  &0.01  &1  &0.08  &0.31  &0.78  \\\cline{2-14}
    & \makecell[l]{TravelTime-387} &0.6  &0.01  &0.33  &0.004  &0.62  &0.07 &0.2  &0.004  &1  &0.004  &0.22  &0.34  \\\cline{2-14}
    & \makecell[l]{TravelTime-451} &1  &0.005  &0  &0  &0  &0 &0  &0  &1  &0.009  &0.82  &0.11  \\
    \hline
    \hline
        \multirow{2}{*}{\makecell[l]{Real Ad\\ Exchange}} & \makecell[l]{exchange-2-cpc-results} & 0.5 & 0.006  & 0 & 0 & 0 & 0 & 0 & 0 & 0.03 & 0.33 & 0.07 & 0.02 \\\cline{2-14}
    & \makecell[l]{exchange-3-cpc-results} & 0.75 & 0.02  & 1 & 0.007 & 0 & 0 & 1 & 0.02 & 0.71 & 0.03 & 0.21 & 0.23 \\
    \hline
    \hline
    \multirow{2}{*}{\makecell[l]{Real Tweets}} & \makecell[l]{Twitter-volume-GOOG} & 0.75 &0.002   &0.38  &0.005 &0.59  &0.02 &0.81  &0.01  &0.75  &0.01  &0.25  &0.43  \\
    & \makecell[l]{Twitter-volume-IBM} &0.37  &0.002  &0.22  &0.005 &0.22  &0.01  &0.5  &0.009  &0.5  &0.005 &0.24 &0.28  \\
    \hline
\end{tabular}
}
\label{Table:ResultsNABDatasets}
\end{table*}
\end{landscape}

\begin{table}[h!t]
\centering
\caption{Optimal window sizes and anomaly classification factors used by OeSNN-UAD for data files from Table~\ref{Table:ResultsNABDatasets}.}
\begin{tabular}{ll|ll}
	\hline
    \multicolumn{2}{c|}{\multirow{1}{*}{Time series}} &  $\mathcal{W}_{size}$ & $\varepsilon$ \\
    \hline\hline
    \multirow{2}{*}{\makecell[l]{Real Ad\\ Exchange}} & \makecell[l]{exchange-2-cpc-results} & 100 & 2 \\\cline{2-4}
    & \makecell[l]{exchange-3-cpc-results} & 100 & 4  \\
    \hline
    \hline
    \multirow{2}{*}{\makecell[l]{Real AWS \\Cloud Watch}} & \makecell[l]{ec2-cpu-utilization-5f5533} & 300 & 2 \\\cline{2-4}
    & \makecell[l]{rds-cpu-utilization-cc0c53} & 600 & 7 \\
    \hline
    \hline
    \multirow{11}{*}{\makecell[l]{Real Known\\ Cause}} & \makecell[l]{ambient-temperature-\\system-failure} &  500 & 6  \\\cline{2-4}
    & \makecell[l]{cpu-utilization-asg-\\misconfiguration}& 600 & 3 \\\cline{2-4}
    & \makecell[l]{ec2-request-latency-\\system-failure} & 400  & 5   \\\cline{2-4}
    & \makecell[l]{machine-temperature-\\system-failure} & 300  & 4  \\\cline{2-4}
    & \makecell[l]{nyc-taxi} &100  &3 \\\cline{2-4}
    & \makecell[l]{rouge-agent-key-hold} & 100  &5 \\\cline{2-4}
    & \makecell[l]{rouge-agent-key-updown} &300  &6  \\
      \hline
      \hline
    \multirow{7}{*}{\makecell[l]{Real Traffic}}
    & \makecell[l]{occupancy-6005} &300  &2   \\ \cline{2-4}
    & \makecell[l]{occupancy-t4013} &600 &2 \\\cline{2-4}
    & \makecell[l]{speed-6005} &600  &2  \\\cline{2-4}
    & \makecell[l]{speed-7578} &100  &4 \\\cline{2-4}
    & \makecell[l]{speed-t4013} &400  &3 \\\cline{2-4}
    & \makecell[l]{TravelTime-387} &100  &2  \\\cline{2-4}
    & \makecell[l]{TravelTime-451} &100  &5  \\
    \hline
    \hline
    \multirow{2}{*}{\makecell[l]{Real Tweets}} & \makecell[l]{Twitter-volume-GOOG} & 200 &3  \\
    & \makecell[l]{Twitter-volume-IBM} & 100  & 5 \\
    \hline
\end{tabular}
\label{Table:NABDatasetsParameters}
\end{table}

\begin{landscape}
\begin{table*}[h!t]
\centering
\caption{Average F-measure values obtained for Yahoo Anomaly Dataset stream data using the unsupervised anomaly detectors (marked with *) presented in \citep{Munir2019_IEEEAccess_DEEPAnT} and using our proposed OeSNN-UAD detector. The bolded results are the best for each data files category. The results for the detectors marked with * were reported in  \citep{Munir2019_IEEEAccess_DEEPAnT}.}
\small{
\begin{tabular}{llllllll}
	\toprule
     Dataset category& Yahoo EGADS* & \makecell[l]{Twitter Anomaly\\ Detection, $\alpha=0.05$*} & \makecell[l]{Twitter Anomaly\\ Detection, $\alpha=0.1$*} & \makecell[l]{Twitter Anomaly\\ Detection, $\alpha=0.2$*} & \makecell[l]{DeepAnT \\ (CNN)*} & \makecell[l]{DeepAnT\\ (LSTM)*} & OeSNN-UAD \\
    \midrule
     A1Benchmark&0.47 &0.48 &0.48 &0.47 &0.46 &0.44 & \textbf{0.70}\\
     \midrule
     A2Benchmark&0.58 &0 &0 &0 &0.94 &\textbf{0.97} & 0.69\\
     \midrule
     A3Benchmark&0.48 &0.26 &0.27 &0.3 &\textbf{0.87} &0.72 & 0.41 \\
     \midrule
     A4Benchmark&0.29 &0.31 &0.33 &0.34 &\textbf{0.68} &0.59 & 0.34\\
	\bottomrule
\end{tabular}
}
\label{Table:ResultsYahoo}
\end{table*}
\end{landscape}

In Table~\ref{Table:NABOtherMeasures}, we provide average values of precision, recall, F-measure (F1), balanced accuracy (BA) and Matthews correlation coefficient (MCC), obtained with OeSNN-UAD for all categories of data files in the NAB and Yahoo repositories.

\begin{table*}[h!t]
\centering
\caption{Average values of F-measure, balanced accuracy (BA) and Matthews correlation coefficient (MCC) obtained with OeSNN-UAD for the NAB and Yahoo repositories.}
\begin{tabular}{llllllll}
	\toprule
    \multicolumn{2}{l}{{Dataset category}}& {Prec.} & {Rec.}  & {F1} & {BA} & {MCC} \\
    \midrule
%   \multirow{8}{*}{\makecell[l]{Numenta \\ Anomaly \\ Benchmark}}  & Artificial no Anomaly& 0 & 0 & 0 & 0 \\
%      \cmidrule{2-6}
    \multirow{7}{*}{\makecell[l]{Numenta \\ Anomaly \\ Benchmark}} & Artificial with Anomaly& 0.500 & 0.457 &0.427 & 0.690 & 0.391  \\
     \cmidrule{2-7}
    & Real Ad Exchange & 0.224 & 0.255 & 0.234 & 0.584 & 0.154  \\
     \cmidrule{2-7}
    & Real AWS Cloud & 0.358 & 0.445 & 0.342 & 0.683 & 0.369  \\
     \cmidrule{2-7}
    & Real Known Cause& 0.263 & 0.469 & 0.324&  0.652&  0.244  \\
    \cmidrule{2-7}
    & Real Traffic& 0.433 & 0.387 & 0.340&  0.646 & 0.299 \\
     \cmidrule{2-7}
    & Real Tweets & 0.267 & 0.412 & 0.310 & 0.633 & 0.225 \\
	\midrule
	\midrule
	\multirow{5}{*}{\makecell[l]{Yahoo \\ Anomaly \\ Dataset}}  & A1Benchmark & 0.657 & 0.791 & 0.697 & 0.869 & 0.706 \\
     \cmidrule{2-7}
    & A2Benchmark & 0.616 & 0.929 & 0.690 &  0.957 & 0.715  \\
     \cmidrule{2-7}
    & A3Benchmark & 0.557 & 0.374 & 0.409 &  0.686 & 0.432  \\
     \cmidrule{2-7}
    & A4Benchmark  & 0.467 & 0.373 & 0.342 &  0.683 & 0.369   \\
    \bottomrule
\end{tabular}
\label{Table:NABOtherMeasures}
\end{table*}

Table~\ref{Table:ResultsZhang} presents a comparison of precision, recall, F-measure, balanced accuracy and MCC for OeSNN-UAD and the unsupervised anomaly detection method offered in \citep{Zhang2019-AnomalyDetectionSliding} for the data files in Real Known Cause and Real Tweets categories of the NAB repository which were used there for validation.
The results presented in Table~\ref{Table:ResultsZhang} were derived from the information provided in \citep{Zhang2019-AnomalyDetectionSliding} about the number of true positives and false positives that were discovered by the anomaly detector offered there and based on the number of anomalous and non-anomalous input values in data files of the NAB repository.

\begin{landscape}
\begin{table*}[h!t]
\centering
\caption{Precision, recall, F-measure (F1), balanced accuracy, and Matthews correlation coefficient obtained for OeSNN-UAD and the method of \citep{Zhang2019-AnomalyDetectionSliding} for the selected data files in the NAB repository. The results for the latter method were derived from the information provided in \citep{Zhang2019-AnomalyDetectionSliding}  about the number of true positives and false positives that were discovered by their anomaly detector and based on the number of anomalous and non-anomalous input values in the NAB data files.}
%are calculated based on the number of true positives and false positives that were discovered by their anomaly detector and based on the number of anomalous and non-anomalous input values in data files of the NAB repository.
\small{
\begin{tabular}{ll|lllll|lllll}
	\hline
    \multicolumn{2}{c|}{\multirow{2}{*}{Time series}} & \multicolumn{5}{l|}{\makecell[c]{The method of \citep{Zhang2019-AnomalyDetectionSliding}}} &   \multicolumn{5}{l}{\makecell[c]{OeSNN-UAD}} \\
    & & Prec. & Rec. & F1 &  BA & MCC & Prec.& Rec. & F1 & BA & MCC \\
    \hline\hline
    \multirow{11}{*}{\makecell[l]{Real \\ Known\\ Cause}} & \makecell[l]{ambient-temperature-\\system-failure}& 0.833 &     0.007 &     0.014 &           0.503 &     0.07 & 0.207 &     0.752 &     0.325 &          0.716 &     0.27   \\ \cline{2-12}
    & \makecell[l]{cpu-utilization-asg-\\misconfiguration}&0.5   &     0.001 &     0.003 &        0.501 &     0.022 & 0.318 &     0.494 &     0.387 &         0.699 &     0.328 \\\cline{2-12}
    & \makecell[l]{ec2-request-latency-\\system-failure} &1     &     0.006 &     0.011 &         0.503 &     0.073 & 0.38  &     0.402 &     0.39  &          0.67  &     0.332 \\\cline{2-12}
    & \makecell[l]{machine-temperature-\\system-failure} &0.294 &     0.002 &     0.004 &         0.501 &     0.018&  0.395 &     0.5   &     0.441 &          0.707 &     0.374\\\cline{2-12}
    & \makecell[l]{nyc-taxi} &0.722 &     0.013 &     0.025 &          0.506 &     0.087&  0.166 &     0.471 &     0.245 &         0.604 &     0.138 \\\cline{2-12}
    & \makecell[l]{rouge-agent-key-hold} &0.667 &     0.011 &     0.021 &          0.505 &     0.075 & 0.126 &     0.232 &     0.164 &        0.526 &     0.04  \\\cline{2-12}
    & \makecell[l]{rouge-agent-key-updown} &0.667 &     0.008 &     0.015 &         0.504 &     0.064 & 0.251 &     0.432 &     0.317 &       0.645 &     0.23  \\
      \hline
      \hline
    \multirow{10}{*}{\makecell[l]{Real \\ Tweets}} & \makecell[l]{TV-AAPL} & 0.333 &     0.004 &     0.007 &     0.501 &     0.026 &     0.454 &     0.491 &     0.472 &     0.713 &     0.411  \\  \cline{2-12}
     & \makecell[l]{TV-AMZN} & 0.233 &     0.004 &     0.009 &          0.501 &     0.019 &     0.17  &     0.41  &     0.24  &     0.594 &     0.132  \\ \cline{2-12}
     & \makecell[l]{TV-CRM} & 0.615 &     0.005 &     0.01  &          0.502 &     0.049 &     0.313 &     0.551 &     0.399 &        0.708 &     0.328  \\ \cline{2-12}
     & \makecell[l]{TV-CVS} & 0.6   &     0.002 &     0.004 &          0.501 &     0.03  &     0.201 &     0.543 &     0.293 &        0.656 &     0.21   \\ \cline{2-12}
     & \makecell[l]{TV-FB} & 0.375 &     0.002 &     0.004 &         0.501 &     0.021 &     0.261 &     0.18  &     0.213 &         0.562 &     0.146  \\ \cline{2-12}
     & \makecell[l]{TV-GOOG} & 0.4   &     0.003 &     0.006 &        0.501 &     0.027 &     0.248 &     0.429 &     0.314 &          0.65  &     0.236  \\ \cline{2-12}
     & \makecell[l]{TV-IBM} & 0.429 &     0.002 &     0.004 &          0.501 &     0.023 &     0.241 &     0.284 &     0.261 &          0.592 &     0.172   \\ \cline{2-12}
     & \makecell[l]{TV-KO} & 0.286 &     0.004 &     0.007 &      0.501 &     0.023 &     0.132 &     0.339 &     0.19  &          0.546 &     0.063   \\ \cline{2-12}
     & \makecell[l]{TV-PFE} & 0.25  &     0.007 &     0.013 &        0.502 &     0.026 &     0.188 &     0.463 &     0.267 &         0.62  &     0.168   \\ \cline{2-12}
     & \makecell[l]{TV-UPS} &0.389 &     0.009 &     0.017 &          0.504 &     0.046 &     0.463 &     0.428 &     0.445 &          0.687 &     0.387  \\
    \hline
\end{tabular}
}
\label{Table:ResultsZhang}
\end{table*}
\end{landscape}

\subsection{Experiments with Different Values of Window Size and Anomaly Classification Factor}
\label{subsec:ExperimentsWindowACFactor}

\begin{figure*}[h!t]
	\centering
	\includegraphics[width=1.0\linewidth]{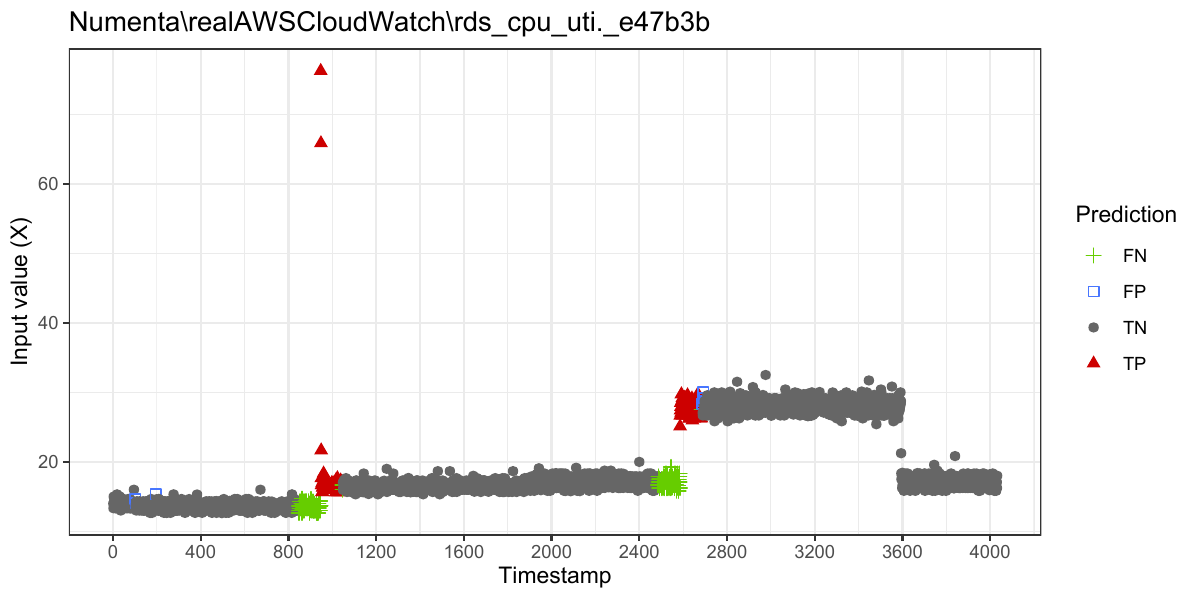}
	\caption{Anomaly detection results for data file \texttt{rds\_cpu\_uti.e47b3b} (the results were obtained with parameters $\mathcal{W}_{size} = 100, \varepsilon = 6$.}
	\label{Fig:rdscpu}
\end{figure*}

In this subsection, we analyse the impact of different values of window size $\mathcal{W}_{size}$ and classification factor $\varepsilon$ on the anomaly detection results obtained by OeSNN-UAD for the data files that were presented in Figures~\ref{Fig:PlotsPredN} and~\ref{Fig:PlotsPredY}, as well as the \texttt{rds\_cpu\_uti.e47b3b} data file (see Figure~\ref{Fig:rdscpu}) from the NAB repository. The first six examined data files contain anomalies of different nature, as discussed in subsection~\ref{subsec:Results}, while the last data file contains a time series covering four data trends and anomalies. In these experiments, the values of all parameters of OeSNN-UAD, except for $\mathcal{W}_{size}$ and $\varepsilon$, are set as given in subsection~\ref{subsec:experimentalsetup}.

\begin{figure*}[h!t]
	\centering
	\includegraphics[width=1.0\linewidth]{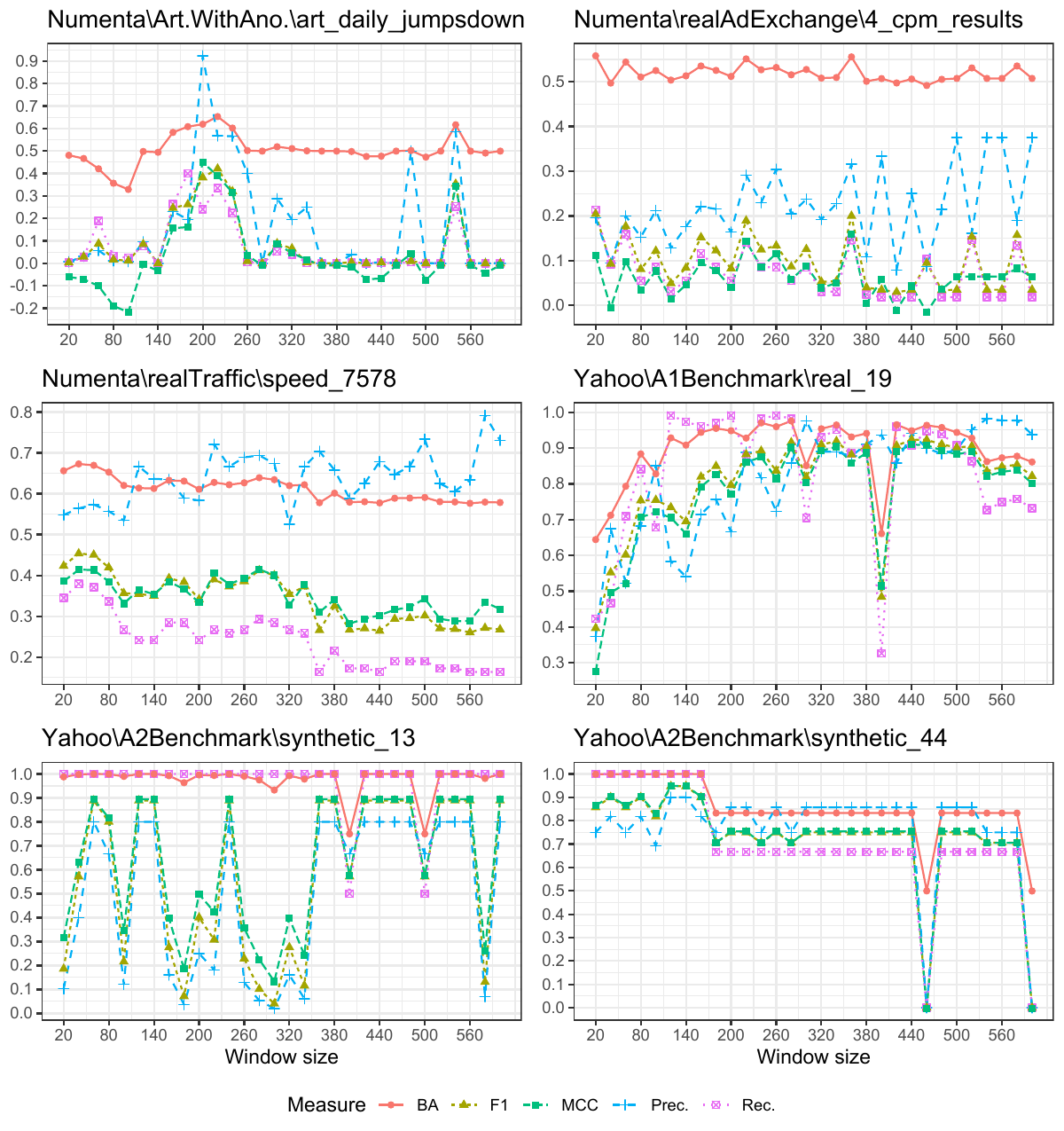}
	\caption{The plots of the obtained measures (precision, recall, F1, BA and MCC) for anomaly detection experiments with OeSNN-UAD using different window sizes $\mathcal{W}_{size}$ for data files presented in Figures~\ref{Fig:PlotsPredN} and~\ref{Fig:PlotsPredY}. The applied values of $\varepsilon$ for data files: \texttt{art\_daily\_jumpsdown}, \texttt{4\_cpm\_results}, \texttt{speed\_7578}, \texttt{real\_19}, \texttt{synthetic\_13} and \texttt{synthetic\_44} are as follows: 5, 3, 4, 4, 4 and 7, respectively.}
	\label{Fig:WindowSizes1}
\end{figure*}

\begin{figure*}[h!t]
	\centering
	\includegraphics[width=1.0\linewidth]{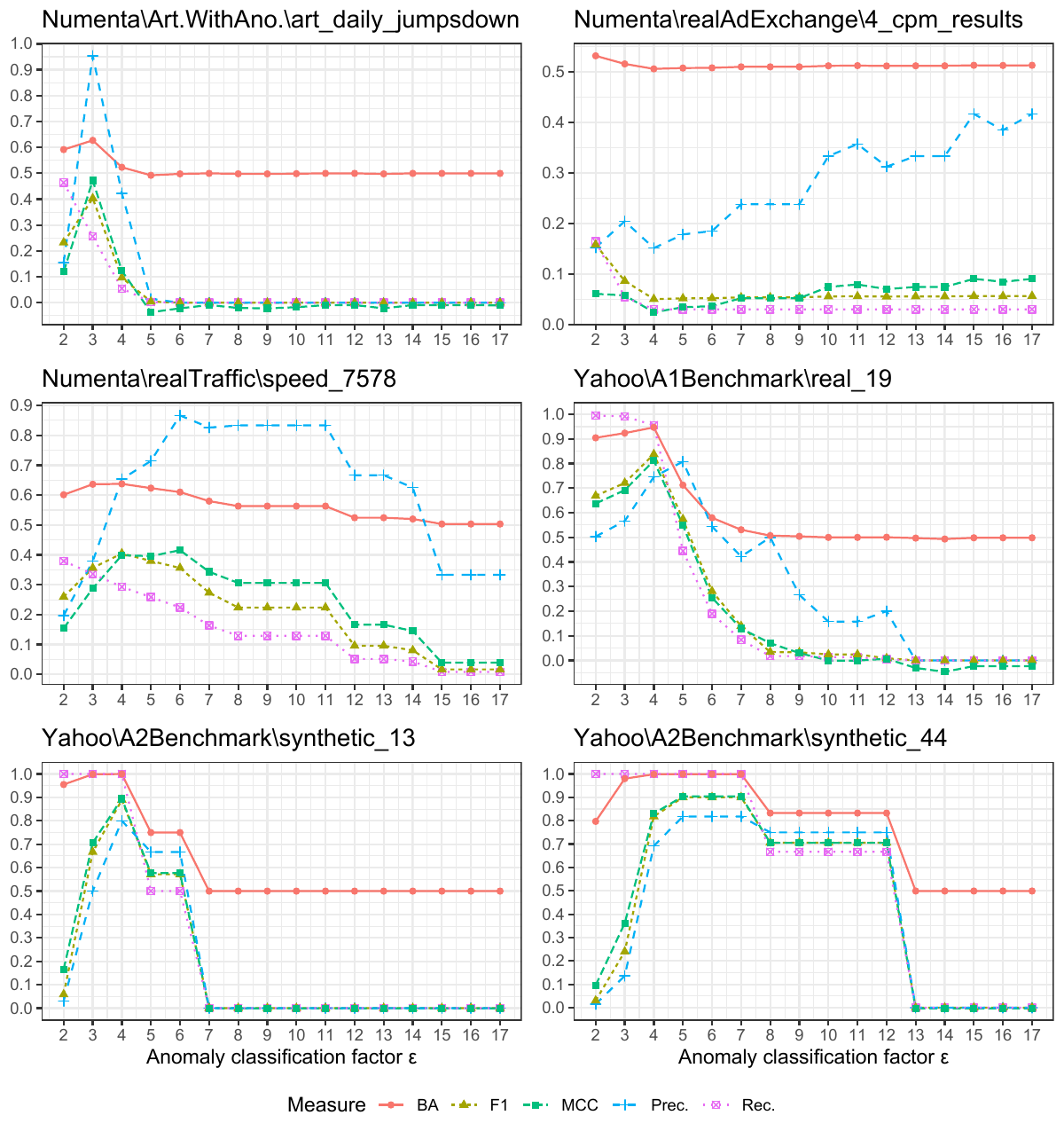}
	\caption{The plots of obtained measures (precision, recall, F1, BA and MCC) for anomaly detection experiments with OeSNN-UAD using different values of anomaly classification factor $\varepsilon$ for data files presented in Figures~\ref{Fig:PlotsPredN} and~\ref{Fig:PlotsPredY}. The applied values of $\mathcal{W}_{size}$ for data files: \texttt{art\_daily\_jumpsdown}, \texttt{4\_cpm\_results}, \texttt{speed\_7578}, \texttt{real\_19}, \texttt{synthetic\_13} and \texttt{synthetic\_44} are as follows: 300, 300, 100, 180, 480 and 80, respectively.}
	\label{Fig:AnomalyFactor}
\end{figure*}

In Figure~\ref{Fig:WindowSizes1}, we present the example plots of the obtained values for the selected performance measures (precision, recall, F1, BA and MCC) for anomaly detection using different values of the window size $\mathcal{W}_{size}$ for data files presented in Figures~\ref{Fig:PlotsPredN} and~\ref{Fig:PlotsPredY}. Analogous plots for different values of the $\varepsilon$ parameter are presented in Figure~\ref{Fig:AnomalyFactor}.

Data files \texttt{art.\_daily\_jumpsdown} (Figure~\ref{Fig:PlotsPredN}), \texttt{synthetic\_13} (Figure~\ref{Fig:PlotsPredY}) and \texttt{synthetic\_44} (Figure~\ref{Fig:PlotsPredY}) can be conceived as periodic time series with a few anomalies. All selected anomaly detection measures obtain optimal values of $\mathcal{W}_{size}$ values approximately equal to:
\begin{itemize}
\item 1 $\times$ the time series period length in the case of \texttt{art.\_daily\_jumpsdown},
\item 0.5 $\times$ the time series period length in the case of data file \texttt{synthetic\_44},
\item \textit{m} $\times$ the time series period length, where $m \in \{0.5, 1, 2, 3\}$, and for some even greater values in the case of data file \texttt{synthetic\_13}.
\end{itemize}

In the case of data files \texttt{real\_19} (Figure~\ref{Fig:PlotsPredY}) and \texttt{speed\_7578} (Figure~\ref{Fig:PlotsPredN}), in which collective anomalies occur, the best results of our anomaly detection approach were obtained for $\mathcal{W}_{size}$ not less than the cardinality of the most numerous maximal cluster of collective anomalies (see Figure~\ref{Fig:WindowSizes1}). In particular, data file \texttt{real\_19} contains three maximal clusters of collective anomalies (see Figure~\ref{Fig:PlotsPredY}), the first of which is most numerous. Thus, $\mathcal{W}_{size}$ is recommended to be set to the cardinality of this cluster for data file \texttt{real\_19}.

In the case of data file \texttt{exchange\_4\_cpm\_results} (Figure~\ref{Fig:PlotsPredN}), multiple values of $\mathcal{W}_{size}$ allowed obtaining comparably good quality measures (see Figure~\ref{Fig:WindowSizes1}). Unlike the data files discussed earlier in this subsection, \texttt{exchange\_4\_cpm\_results} contains non-periodic time series with mainly single point anomalies rather than collective anomalies. The proper determination of the value of $\mathcal{W}_{size}$ for this data file is challenging, especially due to the fact that the current labeling of what is an anomaly and what is not happens to be counter-intuitive (this labeling issue was previously discussed in subsections~\ref{subsec:RepositoriesNAB} and~\ref{subsec:Results}).

In Figure~\ref{Fig:AnomalyFactor}, we present the plots of obtained values of the selected quality measures when only the anomaly classification factor $\varepsilon$ is subject to changes. As it can be noticed from the plots in this figure, the best values of the selected measures are usually obtained for smaller values of $\varepsilon$, such as 3, 4 or 5. However, in the case of \texttt{4\_cpm\_results} data file, the best values of the precision and MCC measures are obtained for larger values of the $\varepsilon$ parameter (the best values of the other measures in the case of this data file were obtained for $\varepsilon = 2$).

\begin{figure*}[h!t]
	\centering
	\includegraphics[width=1.0\linewidth]{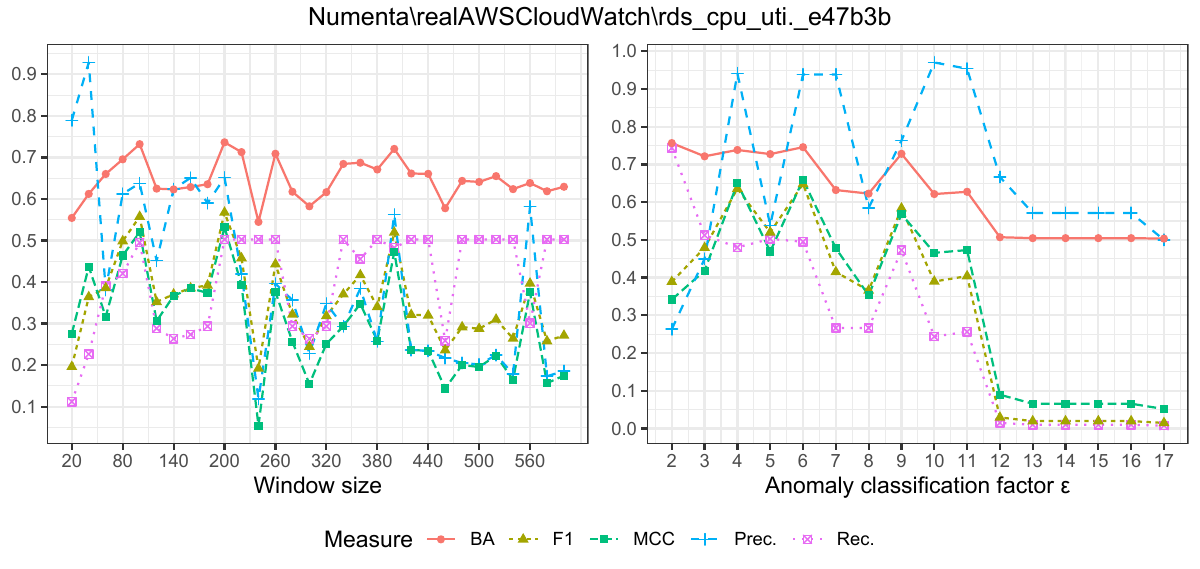}
	\caption{The plots of obtained measures (precision, recall, F1, BA and MCC) for anomaly detection experiments with OeSNN-UAD using different values of window size $\mathcal{W}_{size}$ (left plot) and anomaly classification factor $\varepsilon$ (right plot) obtained for data file \texttt{rds\_cpu\_util\_e47b3b}. The left plot was obtained using constant value of $\varepsilon = 4$, while the right plot was obtained with constant value of $\mathcal{W}_{size} = 180$.}
	\label{Fig:Rds-cpu-window-factor}
\end{figure*}

Furthermore, in order to better assess the impact of $\mathcal{W}_{size}$ and $\varepsilon$ on the anomaly detection results, we carried out experiments on \texttt{rds\_cpu\_uti.e47b3b}, which consists of four data trends and anomalies (see Figure~\ref{Fig:rdscpu}). Please note that input values marked as false negatives at the end of the first two trends presented in this figure are not proper anomalies, but were labeled by the NAB authors as such. Thus, the values of recall for anomaly detection using this data file are understated. Nevertheless, the first input values corresponding to newly observed trends are correctly recognized as anomalies.

The obtained values of anomaly detection quality measures for data file \texttt{rds\_cpu\_uti.e47b3b} and for different values of the $\mathcal{W}_{size}$ and $\varepsilon$ parameters are presented in Figure~\ref{Fig:Rds-cpu-window-factor}. We note that the best values of the selected measures were obtained in particular for $\mathcal{W}_{size}$ equal to 100, which corresponds to the number of input values labeled as anomalous that are present at the beginning of each new trend. In the case of $\varepsilon$, the best values for precision, F1 and MCC as well as values close to the best ones for the recall and BA measures were obtained for both $\varepsilon = 4$ and $\varepsilon = 6$.

The experiments conducted in this subsection suggest that: (i) in the case of a periodic time series, the best anomaly detection results are likely to be obtained for $\mathcal{W}_{size}$ parameter equal to a multiple of the period length of input values or of its half; (ii) in the case of a data file containing different data trends, it seems reasonable to set the value of $\mathcal{W}_{size}$ to the maximum number of consecutive atypical input values that alone are not treated as constituting a new trend in data, but as anomalies;
%the number of input values classified as anomalies after a new trend is recognized corresponds to $\mathcal{W}_{size}$ value;
(iii) in the remaining considered types of data files, it seems reasonable to set the value of $\mathcal{W}_{size}$ to not less than the cardinality of the most numerous maximal cluster of collective anomalies; (iv) the best anomaly detection results can be obtained for relatively small values of the $\varepsilon$ parameter.

%(v) based on the experiments conducted by us, we conclude that the smaller values of $\varepsilon$ can give better anomaly detection results for input values characterized by smaller dispersion, while greater values of this parameter should be usually used with input values characterized by greater dispersion.

%\newpage
\section{Conclusions}
\label{sec:Conclusions and Future Work}

In this article, we offered a new detector of anomalies in data streams. Our proposed  OeSNN-UAD detector is designed for univariate stream time series data and adapts Online evolving Spiking Neural Networks OeSNN. The distinctive feature of our proposed OeSNN-UAD anomaly detector is that it is the only detector based on (Online) evolving Spiking Neural Networks which operates in an online and unsupervised way.

OeSNN-UAD adapts the architecture of OeSNN for anomaly detection purposes. Nevertheless, unlike OeSNN, OeSNN-UAD applies a different model of an output layer and different methods of learning and input values classification. In particular, OeSNN-UAD does not separate output neurons into known in advance decision classes. Instead, each new output neuron on OeSNN-UAD is assigned an output value, which is randomly generated based on recent input values and then is updated in the course of learning of OeSNN-UAD to better adapt to changes in a data stream. As a part of the proposed OeSNN-UAD detector, we offered a new two-step anomaly classification method. Our method treats an input value as anomalous if either none of output neurons fires or, otherwise, if an error between the input value and its OeSNN-UAD prediction is greater than the average prediction error plus user-given multiplicity of the standard deviation of recent prediction errors.

In the article, we proved that all candidate output neurons, as well as all output neurons in the repository, have the same values of the sum of their synaptic weights, their maximal post-synaptic potentials, and their post-synaptic potential thresholds, respectively. The last property eliminates the necessity of recalculation of these thresholds when output neurons of OeSNN-UAD are updated in the course of the learning process, and thus it allows increasing the speed of classification of input stream data. Moreover, we also proved that firing order values of input neurons do not depend on values of $TS$ and $\beta$ parameters, which were previously used in OeSNNs for input value encoding with Gaussian Receptive Fields.

In the experimental part, we compared the quality of the proposed OeSNN-UAD detector with 14 other anomaly detectors provided in the literature. The experiments were conducted on data files from two anomaly benchmark repositories: Numenta Anomaly Benchmark and Yahoo Anomaly Dataset. These two repositories cumulatively contain more than 500 data files grouped into several categories. For the assessment of the quality of anomaly detectors, we used five indicators: F-measure, precision, recall, balanced accuracy and Matthews correlation coefficient. For the Numenta Anomaly Benchmark repository, OeSNN-UAD is able to provide significantly better results in terms of F-measure for all categories of data files. The detailed analysis of the experimental results obtained for the data files in the Numenta Anomaly Benchmark repository that were considered in \citep{Munir2019_IEEEAccess_DEEPAnT} also shows that OeSNN-UAD outperforms other  compared detectors in terms of recall. In the case of the Yahoo Anomaly Dataset repository, OeSNN-UAD achieves higher F-measure values for the real data files category, while for  the other three  synthetic data categories the obtained values of the F-measure are competitive to the results reported in the literature \citep{Munir2019_IEEEAccess_DEEPAnT}. In addition, in terms of recall, F-measure, balanced accuracy and MCC measures, OeSNN-UAD outperforms the method proposed in \citep{Zhang2019-AnomalyDetectionSliding} on all NAB data files which were used there for evaluation.

As we discuss in subsection \ref{subsec:ExperimentsWindowACFactor}, the quality measures of the OeSNN-UAD algorithm depend on properly selected values of its parameters, in particular, the window size $\mathcal{W}_{size}$ and classification factor $\varepsilon$. An inaccurate selection of these parameters can negatively affect the anomaly detection quality. Especially, too large values of the $\varepsilon$ parameter could result in an increase in false negatives (which would result in low recall scores). Conversely, too small values of the $\varepsilon$ parameter could increase the number of false positives (which would entail low precision values). Thus, the value of $\varepsilon$ should be adjusted according to the observed dispersion of at least some representative subset of input values. The selection of proper values for $\mathcal{W}_{size}$ can be even more challenging, as shown in subsection~\ref{subsec:ExperimentsWindowACFactor}. The best value of this parameter can be linked to such factors as the periodicity of time series data, trends present in input values or the cardinalities of collective anomalies clusters. Thus, it is critical to conduct a proper tuning of this parameter that takes into account available characteristics of the input data file under study.

In the performed experiments, we used the OeSNN-UAD detector whose Online evolving Spiking Neural Network contained as few as 10 input neurons and at most 50 output neurons, and thus occupied very little operating memory. In spite of such a small number of neurons, OeSNN-UAD was able to outperform the compared detectors for most data files in each category of the Numenta Anomaly Benchmark repository  and  most real data files in the Yahoo Anomaly Dataset repository. This proves that OeSNN-UAD is effective and suitable also for environments with restrictive memory limits.

\section*{Acknowledgments}
\addcontentsline{toc}{section}{Acknowledgments}

P. Maciąg acknowledges financial support of the Faculty of the Electronics and Information Technology of the Warsaw University of Technology (grant no. II/2019/GD/1). J. L. Lobo and J. Del Ser would like to thank the Basque Government for their support through the ELKARTEK and EMAITEK funding programs. J. Del Ser also acknowledges funding support from the Consolidated Research Group MATHMODE (IT1294-19) given by the Department of Education of the Basque Government.

\bibliographystyle{model5-names}
\biboptions{authoryear}
\bibliography{bib}
\vfill

\end{document}